\documentclass{article}

\usepackage{graphicx}%
\usepackage{multirow}%
\usepackage{makecell}
\usepackage{amsmath,amssymb,amsfonts}%
\usepackage{amsthm}%
\usepackage{mathrsfs}%
\usepackage{xcolor}%
\usepackage{textcomp}%
\usepackage{manyfoot}%
\usepackage{booktabs}%
\usepackage{algorithm}%
\usepackage{algorithmicx}%
\usepackage{algpseudocode}%
\usepackage{listings}%
\usepackage{array}
\usepackage{caption}
\usepackage{pdflscape}
\usepackage[numbers]{natbib} 
\usepackage{subcaption}
\usepackage{threeparttable}
\usepackage{hyperref}
\usepackage[draft]{fixme}

\usepackage{mathtools}

\usepackage[margin=1in]{geometry}

\usepackage{bbm}
\usepackage[T1]{fontenc}

\fxsetup{inline,nomargin,theme=color}

\theoremstyle{plain}

\theoremstyle{definition}

\theoremstyle{remark}

\definecolor{todo-color}{HTML}{ff3232}

\def \isarxiv {Arxiv}

\title{Enhancing Semi-supervised Learning with Zero-shot Pseudolabels}
\author{Jichan Chung$^{1}$ and Irene Y. Chen$^{1,2}$}
\date{$^1$University of California, Berkeley, $^2$University of California, San Francisco}

\begin{document}

\maketitle

\begin{abstract}

The high cost of data labeling presents a major barrier to deploying machine learning systems at scale.
Semi-supervised learning (SSL) mitigates this challenge by utilizing unlabeled data alongside limited labeled examples, while the emergence of foundation models (FMs) offers powerful zero-shot capabilities that can further reduce labeling cost.
However, directly fine-tuning large FMs is often impractical in resource-constrained settings, and naïvely using their pseudo-labels for unlabeled data can degrade performance due to its unreliablity or domain mismatch with target task.
In this work, we introduce ZeroMatch, a novel SSL framework that integrates knowledge distillation with consistency-based learning to jointly leverage labeled data, unlabeled data, and pseudo-labels from FMs. ZeroMatch enables training compact student models using only FM inference, making it suitable for low-resource environments such as personal devices with limited compute. Experiments on six vision and language classification benchmarks show that ZeroMatch consistently outperforms standard SSL and zero-shot augmented methods, demonstrating its effectiveness and robustness across a range of foundation model qualities.

\end{abstract}

\section{Introduction}
\label{intro}

The growing scale of machine learning applications has made data labeling costs a critical bottleneck in deploying ML systems~\cite{northcutt1pervasive,sun2017revisiting,shen2024data}. Semi-supervised learning (SSL) addresses this challenge by leveraging unlabeled data alongside limited labeled examples~\cite{tarvainen2017mean}. Traditional SSL approaches like pseudo-labeling and consistency regularization have demonstrated strong performance across domains, particularly in computer vision and natural language processing~\cite{sohn2020fixmatch,laine2022temporal,tarvainen2017mean}.

In parallel, the emergence of foundation models (FMs) has opened new opportunities for reducing reliance on labeled data.
These large-scale pre-trained models exhibit strong zero-shot capabilities, enabling them to generalize to novel tasks without requiring task-specific fine-tuning~\cite{brown2020language,liangholistic}.
To this end, recent efforts have explored integrating foundation models into the SSL pipeline. Proposed strategies include fine-tuning foundation models with labeled and unlabeled data~\cite{shi2023rethinking,zhang2024candidate,ganerasing}, using zero-shot predictions as pseudo-labels~\cite{hegselmann2023tabllm,nam2023semi}, and distilling knowledge from foundation models into smaller student models~\cite{yang2025knowledge,shiselkd,vemulapalli2023knowledge,zhao2023multistage,jiang2023disco}.

We motivate our problem setting with a practical scenario: a user with a personal device, a modest-sized dataset, and limited computational resources (e.g., a single GPU), with access to foundation model inference services. In such settings, direct fine-tuning of large foundation models is infeasible due to high computational costs. Furthermore, naively using pseudo-labels from foundation models to supervise unlabeled data can degrade performance --- particularly when pseudo-labels are inaccurate~\cite{zhu2024doubly} --- which can be common when the foundation model has not encountered data similar to the user data.
Among existing approaches, knowledge distillation-based methods~\cite{zhao2023multistage,vemulapalli2023knowledge,jiang2023disco} offers a promising solution for such resource-constrained settings. However, current methods typically leverage either the teacher’s predictions or the labeled data independently, missing opportunities to combine these complementary supervision sources. Thus, fully leveraging labels, unlabeled data, and pseudo-labels from foundation models in a unified framework remains an underexplored direction.

In this paper, we introduce \textbf{ZeroMatch}, a method that integrates knowledge distillation and semi-supervised learning to jointly supervise a student model using both pseudo-labels from a foundation model and a confidence-based SSL objective. Our approach is based on the insight that the student model can iteratively refine its predictions on unlabeled data by drawing from the strengths of both the teacher (via distillation) and labeled data (via SSL).
We conduct extensive experiments across six datasets in vision and language classification, evaluating ZeroMatch using pseudo-labels generated by both high- and low-quality foundation models. Our results show that ZeroMatch consistently outperforms standard SSL baselines, zero-shot classification, and other zero-shot augmented approaches, highlighting its robustness and practical applicability in low-resource scenarios.

In summary, our key contributions are as follows:
\begin{enumerate}
\item We propose ZeroMatch: a novel knowledge distillation-based semi-supervised learning approach that effectively leverages labels, unlabeled data, and teacher pseudo-labels from foundation models.

\item Our framework trains small models while directly leveraging zero-shot predictions from foundation models as pseudo-labels, enabling training on modest hardware, and eliminating the need for teacher model to be present. This addresses a practical setting where users have limited computational resources but access to foundation model inference services.

\item We provide extensive experiments on 6 datasets across vision and language domains which demonstrate that ZeroMatch outperforms both standard SSL methods and zero-shot augmented approaches.
\end{enumerate}

\begin{figure*}[t!]
    \centering
    \includegraphics[width=0.95\linewidth]{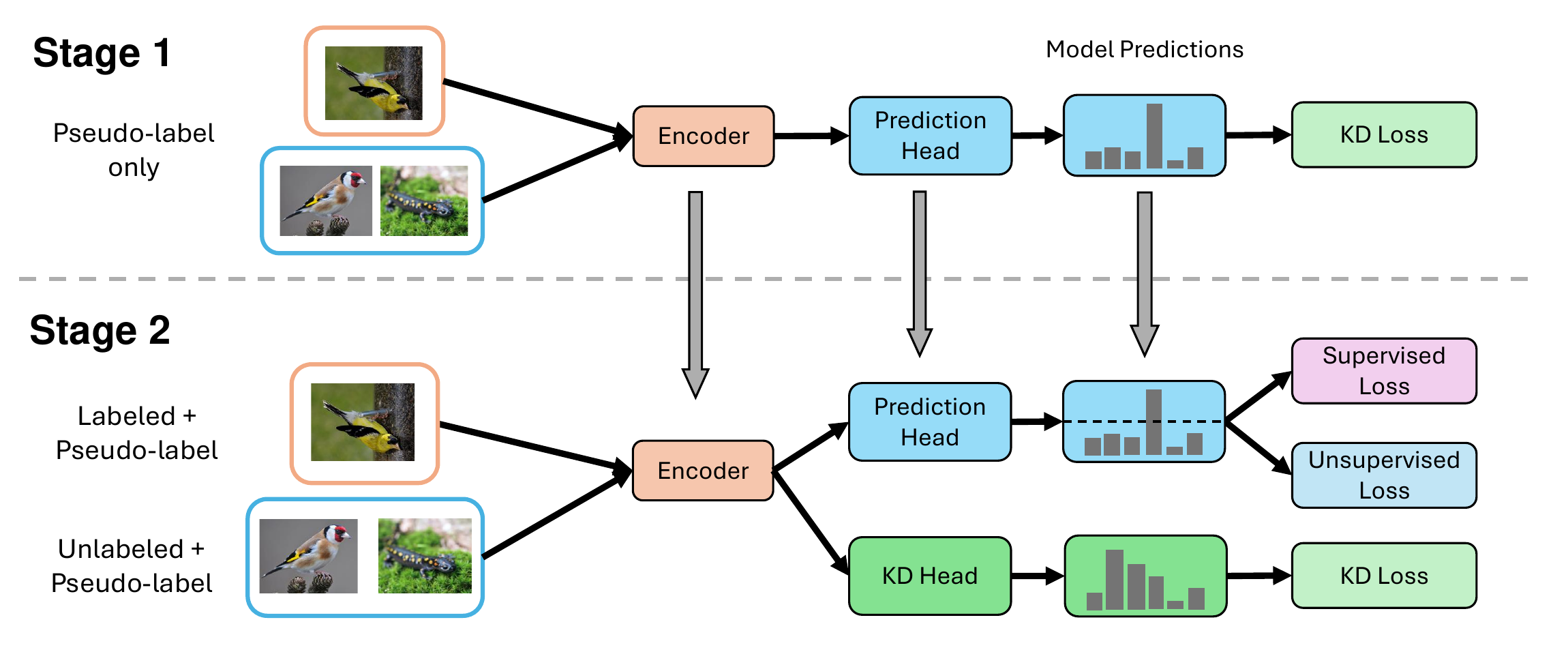}
    \caption{Illustration of our 2-stage ZeroMatch algorithm.
    Both labeled and unlabeled input data receive pseudo-labels from foundation models.
    In stage 1, knowledge distillation is performed with pseudo-labels from teacher foundation model.
    In stage 2, the model is trained with supervised and unsupervised loss of SSL algorithm, with weights and initial confident prediction learned from the previous stage.
    We add an auxiliary classifier head (green box) that runs knowledge distillation task, to reduce potential catastrophic forgetting that may occur during runing SSL.
    }
\label{fig:diagram}
\end{figure*}

\section{Related Work}
\label{sec:related}

\subsection{Semi-supervised Learning}
Semi-supervised learning (SSL) has evolved from foundational consistency regularization methods to sophisticated deep learning approaches. The $\Pi$-Model~\cite{laine2022temporal} and Mean Teacher~\cite{tarvainen2017mean} established core principles by enforcing consistent predictions across different model states. FixMatch~\cite{sohn2020fixmatch} later unified these ideas by combining weak and strong augmentations with pseudo-labeling. Subsequent work focused on reliability: UPS~\cite{rizvedefense} introduced confidence-based filtering while FlexMatch~\cite{zhang2021flexmatch} developed adaptive thresholding for pseudo-label selection. Recent approaches like SimMatch~\cite{zheng2022simmatch} have further advanced the field by incorporating contrastive learning principles.

\subsection{Foundation Models in SSL}
The integration of foundation models into SSL frameworks is an emerging direction.
One of main emerging approaches is to fine-tune foundation models with SSL or unsupervised algorithms~\cite{shi2023rethinking,you2024diffusion,hegselmann2023tabllm,menghini2023enhancing,zhang2024candidate,ganerasing}.
These methods provides a strong and robust learner in general, but are not often applied to state-of-the-art foundaton models due to building upon model-specific features like cosine similarity of embeddings, and requiring significant compute for the fine-tuning step involved in tuning large foundation models.
In contrast, our setup leverage foundation models as a pseudo-label generator.
Researchers have also proposed theoretical guarantees to accommodate potentially noisy pseudo-labels~\cite{zhu2024doubly}. Our work builds on this goal to develop a practical method for incorporating unreliable pseudo-labels.

\subsection{Knowledge Distillation in SSL}
Knowledge distillation (KD) has been widely used to transfer knowledge from large teacher models to smaller student models~\cite{buciluǎ2006model,hinton2015distilling,west2021symbolic,beyer2022knowledge,hsieh2023distilling,voge2024leveraging}, allowing the benefits of large models on low compute device.
Many KD approaches applied in the semi-supervised learning (SSL) setting rely on pre-training a large teacher model—either via supervised or self-supervised objectives~\cite{chen2020big,he2021semi,yang2025knowledge,xie2022spd,shiselkd,vemulapalli2023knowledge} which often makes impractical to apply to recent large foundation models like LLaMA-70B due to its large compute requirement for tuning.
Recent alternative KD attempts avoids explicitly training a teacher by optimizing the student models on teacher's zero-shot outputs, but trains the student model on either one of teacher model outputs~\cite{zhao2023multistage,jiang2023disco} or labeled data~\cite{vemulapalli2023knowledge}, missing synergic opportunity of using both.
To the best of our knowledge, there is no existing KD approach that satisfies our goal to fully leverage labels, unlabeled data and teacher outputs simulataneously.

\section{Preliminaries}
\label{sec:problem-setup}

\subsection{Problem Setup}

Consider a semi-supervised setting with read-only access to a foundation model $\hat{f}$. The training data consists of a labeled set $\mathcal{D}_L = {(x_i, y_i) : i \in [N_L] }$ and an unlabeled set $\mathcal{D}_U = {(u_i) : i \in [N_U] }$, where $\mathcal{D} = \mathcal{D}_L \cup \mathcal{D}_U$ represents the full dataset and $[N]$ denotes integers ${1, 2,..., N}$.
The foundation model $\hat{f}$ generates pseudo-label distributions $\hat{y}^L_i = \hat{f}(x_i)$ for labeled data and $\hat{y}^U_i = \hat{f}(u_i)$ for unlabeled data. 
The goal is to train a classifier $f$ that outputs class distributions $\mathbf{p}(y|x)$ using $\mathcal{D}_L$, $\mathcal{D}_U$, and pseudo-labels $\hat{y}^L_i, \hat{y}^U_i$.
\footnote{Our usage of "pseudo-labels" differs from traditional SSL literature, where they typically refer to predictions on $\mathcal{D}_U$ from a model partially trained on $\mathcal{D}_L$. Here, they represent foundation model predictions without any task-specific training.}
We assume $\hat{f}$ was not trained on $\mathcal{D}_L$ or $\mathcal{D}_U$. The pseudo-label quality depends on the model's capabilities and prompt design, with potential failure modes including hallucinations and out-of-domain responses. 

Our main objective is to develop a SSL method that maximally leverages labels, unlabeled data, and pseudo-labels from foundation model. This implies that the algorithm should fully utilze pseudo-labels when they are high quaility, and gracefully degrades to standard SSL performance when pseudo-labels are unreliable.

\subsection{Motivation: When is Our Setup Most Useful?}

While our setup considers a semi-supervised learning (SSL) scenario where pseudo-labels are generated by foundation models (FMs), in practice, there are many design choices to make when incorporating FMs into SSL workflows. For example:
\begin{itemize}
    \item Choosing between large FMs (LFMs) (e.g., GPT-4o, LLaMA-70B) and smaller FMs (SFMs) (e.g., LLaMA-3B, LLaMA-8B)
    \item Selecting open-weight vs. closed-weight FMs,
    \item Deciding whether to fine-tune FMs or use them in inference-only mode.
\end{itemize}
We clarify where our setup becomes particularly useful.
Assuming that users can access to FM outputs via inference APIs, and may lack access to high-memory compute infrastructure, our setup of using zero-shot pseudo-labels in SSL enables following key benefits:
\begin{enumerate}
\item  \textbf{Our setup can leverage Large FMs for SSL with Limited Compute} -
our setup can involve high quality information of LFMs to SSL task without high computational demands of fine-tuning them, and be often beneficial compared with involving smaller FMs, due to LFM's richer and broader knowledge base.
\item  \textbf{Our setup is Compatible with Closed-weight FMs} - Since our approach only requires textual outputs (e.g., pseudo-labels) from the LFM, it works seamlessly with closed-weight models accessible through inference-only APIs, broadening the choices of FMs to use.
\item  \textbf{Our Setup is Reduces the Risk of Data Leakage.} - By limiting data exposure to a single inference pass and avoiding repeated transmission or remote training, our setup minimizes the risk of sensitive data leak during interaction with remotely hosted LFMs.
\end{enumerate}
We discuss above claims in detail in Appendix~\ref{sec:setup_useful}.
Among the available outputs of foundation models that can be obtained from inference (e.g., embeddings, logits, reasonings), our problem setup focus on zero-shot prompted pseudo-labels for their high performance, simplicity, and human interpretability.
We discuss more details of reasoning for this choice in Appendix~\ref{sub:pseudo_labels_vs_embedding}.

\subsection{Semi-supervised learning methods}
Recent advancements in SSL have led to the development of methods that jointly train a classifier from both supervision with labeled set, and unsupervised feature learning on unlabeled data. The training objective is typically expressed as $\mathcal{L} = \mathcal{L}_s + \mathcal{L}_u$ where $\mathcal{L}_s$ is the supervised loss, often the cross-entropy loss computed on the labeled data: 
\begin{align*}
\mathcal{L}_{s} = \frac{1}{B_L} \sum^{B_L}_{i=1} \mathcal{H}(y_i, \mathbf{p}(y|x_i)). 
\end{align*}
where $B_L$ is the size of data batch sampled from labeled set $\mathcal{D}_L$.

On the other hand, $\mathcal{L}_u$ represents the unsupervised loss, which leverages different strategies to incorporate unlabeled data into the training process. Some notable strategies include the following:
\begin{itemize}
    \item \textbf{Self-training}: This strategy involves using the model trained on labeled data to generate temporary predictions for the unlabeled samples. These predictions are then incorporated into the model's supervision objective~\cite{lee2013pseudolabel}.
    \item \textbf{Confidence-based Sample Selection}: Unlabeled samples with not confident or potentially incorrect pseudo-labels are filtered out based on predefined thresholds. This ensures that only samples with reliable predictions contribute to the training process~\cite{qizhe2020}.
    \item \textbf{Strong Augmentation}: To further enhance feature learning, the model is optimized so that its predictions on weakly augmented and strongly augmented versions of the same sample agree ~\cite{qizhe2020}. This encourages the model to learn robust features.
    \item \textbf{Distribution Alignment}: This technique adjusts the output probability distribution of each class based on the input data distribution~\cite{berthelot2020}. 
\end{itemize}
With strong augmentation $\mathcal{A}_s(u_i)$ and weak augmentation $\mathcal{A}_w(u_i)$ of unlabeled samples $u_i$, we denote predictions from classifier $f$ of these augmented samples by 
$p^s_i=\mathbf{p}(y|\mathcal{A}_s(u_i))$ and $p^w_i=\mathbf{p}(y|\mathcal{A}_w(u_i))$, respectively.

When a data batch of size $B_U$ is sampled from $\mathcal{D}_U$, the unsupervised training objective in these methods typically takes the following form:
\begin{align*}
\mathcal{L}_{u} = \frac{1}{B_U} \sum^{B_U}_{i=1} \mathbbm{1}(\text{max}(\hat{p}^w_i) > \tau) \mathcal{H}(y_i, \mathbf{p}(y|u_i))
\end{align*}
where $\hat{p}^w_i = DA(p^w_i)$ is label prediction for input $\mathcal{A}_w(u_i)$ and $DA$ represents distribution alignment.



\section{The ZeroMatch Framework}
\label{sec:zeromatch}

In order to develop a SSL method that effectively leverages labeled data, unlabeled data, and pseudo-labels generated by a foundation model, we propose a hybrid approach that combines the strengths of knowledge distillation and SSL techniques.

The two key intuitions behind our approach are:
\begin{enumerate}
    \item \textbf{Knowledge distillation methods utilizes teacher model's prediction} to improve student model's prediction output on unlabeled data.
    \item \textbf{Semi-supervised learning algorithms utilize labeled data} to refine model's predictions on unlabeled data.
\end{enumerate}
which implies that both methods can be used to improve predictions on unlabeled data from different sources of supervision in a complementary manner. To this end, we propose following two-stage algorithm that integrates both paradigms to maximize the utility of all available supervision signals.

\paragraph{Stage 1: Knowledge Distillation (KD)}
We start by following standard knowledge distillation procedure with teacher's output.
A student model is trained using available unlabeled data (including the unlabeled input in labeled data) with the pseudo-labels $\hat{y}$ generated from teacher foundation model $\hat{f}$ with following objective:
\begin{align*}
\mathcal{L}_{KD} = \frac{1}{N} \left( \sum^{N_L}_{i=1} \mathcal{H}(\hat{y}^L_i, \mathbf{p}(y|x_i)) + \sum^{N_U}_{i=1} \mathcal{H}(\hat{y}^U_i, \mathbf{p}(y|u_i)) \right).
\end{align*}
where $N=N_L+N_U$.
The resulting student model obtains high-confidence predictions for unlabeled data that matches the teacher's pseudo-labels.

\paragraph{Stage 2: Semi-supervised learning with auxiliary KD loss}

The main objective of second stage is to train the student model with SSL objective.
Unlike running SSL from scratch, predictions obtained from previous stage can be utilized as initial high-confidence samples in SSL algorithm, allowing more unlabeled samples to be utilized since the beginning stage of SSL learning, achieving faster convergence as result.

One possible failure mode of running a SSL algorithm in stage 2 is catastrophic forgetting~\cite{goodfellow2013empirical}, where learned knowledge from teacher's pseudo-label can be forgotten during training for downstream task (SSL in this case).
This can happen frequently in low-label setting, where SSL algorithm can develop inaccurate predictions on unlabeled data due to inherently low information in labels, leading to inaccurate pseudo-label that overwrites the knowledge learned.
To address this issue, we include knowledge distillation as auxiliary objective alongside SSL objective, motivated by~\cite{kar2022preventing}.
When the student classifier $f$ consists of a non-linear projector head $h(\cdot)$ and a feature encoder $g(\cdot)$, such that $f = h \circ g $, we introduce an additional linear projector head $h_p(\cdot)$, which learns a knowledge distillation task, while sharing the encoder $g$ with the main classification task.
Given a batch of labeled and unlabeled data of size $B_L$ and $B_U$, stage 2 optimizes $h_p \circ g$ with the following loss:
\begin{align*}
\mathcal{L}_{KD_2} = \frac{1}{B} \left( \sum^{B_L}_{i=1} \mathcal{H}(\hat{y}^L_i, \mathbf{q}(y|x_i)) + \sum^{B_U}_{i=1} \mathcal{H}(\hat{y}^U_i, \mathbf{q}(y|u_i)) \right).
\end{align*}
where $ B = B_L + B_U$ and $\mathbf{q}(y|x)$ indicates predicted class distribution of $h_p(g(x))$.
Together with the SSL objective, the overall loss function of stage 2 is:
\begin{align*}
\mathcal{L}_{KD-SSL} = \mathcal{L}_s + \mathcal{L}_u + \alpha_t \cdot \lambda_p \mathcal{L}_{KD_2}
\end{align*}
where $\alpha_t$ is annealing parameter that linearly increases value from $0$ to $1$ during training based on training step $t$, and $\lambda_p$ is a fixed scalar hyperparameter indicating relative weight of pseudo-label prediction task. We denote the inclusion of annealing with binary variable $\alpha_p$.
This modification ensures continuous supervision of the teacher's pseudo-label into student's model during the SSL training, mitigating the catastropic forgetting issue.
Our algorithm is illustrated in Figure~\ref{fig:diagram}.

Our algorithm can robustly improve from both labels and pseudo-labels.
When the pseudo-labels are accurate, the model benefits from the knowledge distillation task throughout both stage of training. Conversely, when the pseudo-labels are noisy, the model can still benefit from SSL's objective, with less the impact from the inaccurate knowledge distiallation due to task being learned at the separate classification head $h_p$. We integrate our method with AdaMatch~\cite{berthelotadamatch}, which is a highly performant SSL baseline across various datasets and domains, as demonstrated by benchmark results~\cite{wang2022usb}. We denote this method by \textbf{ZeroMatch}.

\begin{table*}[t!]
    \centering
    \small
    \caption{Summary of data used in the experiments. For each dataset, we run experiments with at most 3 pseudo-label sets from different foundation models. The foundation model $\hat{f}$ used to generate each pseudo-label set are shown below.}
    \begin{tabular}{@{}ccccccc@{}}
    \toprule
    Domain               & Dataset       & \# Train & \# Val & \#Test  & \# Class & Foundation model $\hat{f}$                                                                          \\ \midrule
    \multirow{3}{*}{NLP} & Amazon Review & 250,000  & 25,000 & 650,000 & 5        & \multirow{3}{*}{\begin{tabular}[c]{@{}c@{}}GPT-4o, Llama3.3-70B, \\ FLAN-T5\end{tabular}} \\
                         & AG News       & 100,000  & 10,000 & 7,600   & 4        &                                                                                           \\
                         & Yahoo! Answer & 500,000  & 50,000 & 60,000  & 10       &                                                                                           \\ \midrule
    \multirow{3}{*}{CV}  & CIFAR100      & 50,000   & -      & 10,000  & 100      & \multirow{3}{*}{GPT-4.1, CLIP}                                                            \\
                         & Flowers102    & 1,020    & -      & 6,149   & 102      &                                                                                           \\
                         & Resisc45      & 3,150    & -      & 25,200  & 45       &                                                                                           \\ \bottomrule
    \end{tabular}
    \label{tab:data_summary}
\end{table*}

\section{Experiments}
\label{experiments}

\subsection{Experiment Setup}

\paragraph{Datasets}
We use 6 publicly-available datasets across vision and natural language processing (Table~\ref{tab:data_summary}). Dataset sizes range from 100s to 100,000s of data points.
To evaluate the effect of labeled dataset sizes, for each dataset, we create up to three tasks with different sizes of labeled data while the number of total unlabeled data points stays the same.
When sampling the labeled datapoints, same number of labeled samples are drawn for each class.
For example, in the CIFAR100 dataset, we create sets with 100, 200, and 400 labeled data points, each having 1, 2, 4 labels per class, respectively. The explanation of each dataset is provided in Appendix~\ref{sec:appendix-data}.

\paragraph{Teacher Pseudo-labels from Foundation Models}
For each dataset, we consider multiple sets of pseudo-labels with varying quality.
For NLP datasets, we generate pseudo-labels with GPT-4o~\cite{openai2024gpt4o}, LLama3.3-70B~\cite{meta2024llama33}, and FLAN-T5~\cite{chung2024scaling}. Among FLAN-T5 models with different weight sizes, we choose two models: one with best quality, and one having lowest quality, based on its test accuracy. We name each pseudo-label set as `A' and `B' in the order of decreasing quality. See Appendix~\ref{sec:zero_shot_benchmark} for exact model used.
For image dataset, we use GPT-4.1~\cite{openai2025gpt41} and CLIP~\cite{radford2021learning}.
We provide benchmark results of zero-shot pseudo-labels on test set in Table~\ref{tab:zero_shot_nlp} in Appendix~\ref{sec:zero_shot_benchmark}.
Further information on pseudo-label generation procedure can be found in Appendix~\ref{sec:plabel}.


\paragraph{Training Setup}
Our data-specific training configuration is explained in detail in Appendix~\ref{sec:training-setup}.
For image classification, we train ViT-Small~\cite{dosovitskiy2020image} with AdamW optimizer for $T =$204,800 steps for CIFAR100, and $T = $102,400 steps for Flowers102 and Resisc45.
For NLP classification tasks, we use pre-trained BERT-Base~\cite{devlin2018bert} using the AdamW optimizer for $T =$102,400 steps.
To ensure a fair comparison with AdaMatch (an SSL baseline), we use the exact same hyperparameters used for AdaMatch.
For hyperparmeters specific to our method, we use $\alpha_p = 1$ (indicating that annealing is applied) and $\lambda_p=1$ for all experiments.
Discussion on default choice of $\alpha_p$ and $\lambda_p$, and its sensitivity analysis is provided in Appendix~\ref{sec:hyperparameter_sensitivity}.

\paragraph{Implementation details}
We implement our method using the Unified SSL Benchmark (USB) framework~\cite{wang2022generalizing}.
To minimize the complexity of integrating our method with the existing SSL baseline, we simply add an additional multi-layer perceptron (MLP) classification head with the same architecture as the baseline model's MLP head.
All experiments are conducted in PyTorch with a single NVIDIA A5000 24GB GPU, but we note that our experiments can run on a GPU with much smaller VRAM due to training backbone being small (see Section~\ref{sec:compute_cost} for computational requirements of our method.).

\subsection{Baselines}

\paragraph{Pseudo-supervision}
When given a semi-supervised dataset that includes pseudo-labels, a straightforward approach is to fill in the prediction targets for the unlabeled samples using the pseudo-labels, resulting in fully pseudo-labeled unlabeled set $\hat{\mathcal{D}_U} = \{(u_1, \hat{y}^U_1) , (u_2, \hat{y}^U_2),  \ldots, (u_{N_U}, \hat{y}^U_{N_U})\}$. The model can then be trained using a supervised learning algorithm on this fully-labeled d{}ataset $\hat{\mathcal{D}} = \mathcal{D}_L \cup \hat{\mathcal{D}_U}$ . We refer to this method as \textit{pseudo-supervision}. We refer to Appendix~\ref{sec:imp_detail_baseline} for implementation details.

\paragraph{Pseudo-label as Feature Input}
Another straghtforward approach to use pseudo-label to help in semi-supervised setting is to include them as one-hot features, which is often employed in tabular learning setting to include categorical features~\cite{borisov2022deep}.
We use one-hot converted pseudo-labels as addtional feature input to the model.
This approach can learn to accomodate or ignore pseudo-label feature depending on how helpful they are in learning the task.
We name this approach to be \textit{PL feature input}.
See Appendix~\ref{sec:imp_detail_baseline} for implementation details.


\paragraph{Additional comparison with previous works}
We discuss additional comparisons of ZeroMatch with other previous works in Appendix~\ref{sec:additional_comparsion}.
While our work mainly focus on involving foundation model to SSL setting in pseudo-label format, there are other previous works
outside this category that also tries to enhance SSL with foundation models (ex. GRIP~\cite{menghini2023enhancing}, CPL~\cite{zhang2024candidate} and FineSSL~\cite{ganerasing}).
We provide overview of how these method compares with ZeroMatch in practice in Appendix~\ref{sec:additional_comparsion_ft_fm}.
We also provide additional comparisons to previous works like Doubly-robust self-training~\cite{zhu2024doubly}.

\subsection{ZeroMatch Enhances SSL with High Quality Pseudo-labels from Large Foundation Models}

\begin{table*}[t!]
\centering
\small
\caption{Accuracy (\%) in Yahoo Answers, AG News, Amazon Review with pseudo-labels from GPT-4o and LLama3.3-70B. Median and standard deviation of 3 different random seeds are reported. Best score among experiments with same pseudo-label set are in bold. `FM' refers to the foundation model used to generate the pseudo-label set.}
\resizebox{\textwidth}{!}{
\begin{tabular}{c|c|ccc|ccc|ccc}
\hline
                              & Dataset          & \multicolumn{3}{c|}{Yahoo Answers}                              & \multicolumn{3}{c|}{AG News}                                    & \multicolumn{3}{c}{Amazon Review}                               \\ \hline
FM                             & Label size       & 250                 & 500                 & 2000                & 20                  & 40                  & 200                 & 125                 & 250                 & 1000                \\ \hline
None                          & Adamatch         & 64.81±1.29          & 67.3±0.53           & 69.42±0.42          & 88.13±2.55          & 85.21±1.44          & 88.33±1.94          & 45.03±4.05          & 52.39±1.65          & 56.18±0.45          \\ \hline
\multirow{4}{*}{GPT-4o}       & Zero-shot        & 68.81±0.00          & 68.81±0.00          & 68.81±0.00          & 86.25±0.00          & 86.25±0.00          & 86.25±0.00          & 59.14±0.00          & 59.14±0.00          & 59.14±0.00          \\
                              & Pseudo-supervise & 67.68±0.06          & 67.30±0.20          & 67.56±0.06          & 86.13±0.15          & 86.33±0.08          & 86.33±0.14          & 56.94±0.27          & 56.65±0.21          & 56.83±0.25          \\
                              & PL feature Input & 70.56±0.27          & 71.00±0.57          & 71.47±0.67          & 86.25±0.46          & 88.00±0.90          & 88.39±1.00          & 50.38±0.67          & 53.30±1.23          & 57.08±0.39          \\
                              & ZeroMatch (ours) & \textbf{70.90±1.07} & \textbf{71.11±0.09} & \textbf{72.09±0.32} & \textbf{88.50±0.22} & \textbf{88.70±0.22} & \textbf{88.76±1.20} & \textbf{60.12±0.33} & \textbf{59.82±0.39} & \textbf{60.19±0.48} \\ \hline
\multirow{4}{*}{\begin{tabular}[c]{@{}c@{}}LLama3.3\\ -70B\end{tabular}} & Zero-shot        & 69.15±0.00          & 69.15±0.00          & 69.15±0.00          & 88.41±0.00          & 88.41±0.00          & 88.41±0.00          & 55.79±0.00          & 55.79±0.00          & 55.79±0.00          \\
                              & Pseudo-supervise & 67.85±0.21          & 67.67±0.39          & 67.17±0.14          & 88.47±0.08          & 88.53±0.18          & 88.45±0.08          & 53.74±0.33          & 54.34±0.11          & 54.14±0.53          \\
                              & PL feature input & 70.70±0.23          & 71.37±0.21          & \textbf{72.18±0.25}          & 86.79±7.98          & 88.41±0.02          & 88.24±0.92          & 51.55±0.93          & 54.39±0.79          & 57.01±0.37          \\
                              & ZeroMatch (ours) & \textbf{71.28±0.29} & \textbf{71.46±0.95} & 71.68±0.19 & \textbf{88.79±0.05} & \textbf{88.96±0.07} & \textbf{88.86±0.17} & \textbf{57.36±0.81} & \textbf{59.46±0.13} & \textbf{60.06±0.28} \\ \hline
\end{tabular}
}
\label{tab:nlp_gpt_results}
\end{table*}


\begin{table*}[t!]
\centering
\small
\caption{Accuracy (\%) in CIFAR-100, Flowers102, Resisc45 with pseudo-labels from GPT-4.1. Median and standard deviation of 3 different random seeds are reported. Best score among experiments with same pseudo-label set are in bold. `FM' refers to the foundation model used to generate the pseudo-label set. }
\begin{tabular}{c|c|ccc|c|c}
\hline
                         & Dataset          & \multicolumn{3}{c|}{CIFAR100}                                   & Flowers102          & Resisc45            \\ \hline
FM                        & Label size       & 100                 & 200                 & 400                 & 204                 & 90                  \\ \hline
None                     & Adamatch         & 71.43±2.48          & 78.32±0.46          & 84.02±0.74          & 86.71±0.69          & 78.87±0.80          \\ \hline
\multirow{4}{*}{GPT-4.1} & Zero-shot        & 83.25±0.00          & 83.25±0.00          & 83.25±0.00          & 88.37±0.00          & 79.28±0.00          \\
                         & Pseudo-supervise & 84.84±0.03          & 84.93±0.27          & 85.12±0.13          & 85.40±0.25          & 79.59±0.18          \\
                         & PL feature input & 72.81±1.33          & 82.48±1.87          & 85.78±0.44          & 89.53±1.44          & 82.01±1.09          \\
                         & ZeroMatch (ours) & \textbf{88.01±0.05} & \textbf{87.97±0.20} & \textbf{88.12±0.14} & \textbf{95.17±0.88} & \textbf{87.83±0.72} \\ \hline
\end{tabular}
\label{tab:image_gpt_results}
\end{table*}

Benchmark results for the NLP datasets with GPT-4o and LLama3.3-70B are presented in Table~\ref{tab:nlp_gpt_results}, and image datasets with GPT-4.1 are presented in Table~\ref{tab:image_gpt_results}.
When combined with pseudo-labels from high-quality large foundation models, ZeroMatch achieves the highest scores in most of the settings (22 of 23 total settings, except 2000-label case of Yahoo Answers with LLama3.3-70B).
Our results shows that ZeroMatch largely improves upon SSL baseline and Zero-shot.
For example, in Amazon Review dataset with 40 labels and pseudo-label from LLama3.3-70B, ZeroMatch achieves an $71.28\%$ compared to Zero-shot of $69.15\%$ and Adamatch $64.81\%$, yielding an improvement of $2.13\%$ and $6.47\%$ respectively.
Our method also outperforms pseudo-supervision and PL feature input baselines, with highest improvement coming from
100-label setting of CIFAR100 with GPT-4.1 where ZeroMatch produces $15.2\%$ improvement compared to PL feature input, and
$8.23\%$ improvement in 90-label setting of Resisc45 with GPT-4.1 for pseudo-supervision.

\subsection{ZeroMatch Improves Performance on SSL Baselines Despite Low-Quality Teacher Pseudo-labels}

\begin{table*}[t!]
\centering
\small
\caption{Accuracy (\%) in Yahoo Answers, AG News, Amazon Review with pseudo-label from FLAN-T5 models. `A' and `B' in FM column indicates using highest and lowest quality pseudo-label sets (decreasing in order) from FLAN-T5 models. Median and standard deviation of 3 different random seeds are reported. Best score among experiments with same pseudo-label set are in bold.  }
\resizebox{\textwidth}{!}{
\begin{tabular}{c|c|ccc|ccc|ccc}
\hline
                   & Dataset                & \multicolumn{3}{c|}{Yahoo Answers}                              & \multicolumn{3}{c|}{AG News}                                    & \multicolumn{3}{c}{Amazon Review}                               \\ \hline
FM                 & Label size             & 250                 & 500                 & 2000                & 20                  & 40                  & 200                 & 125                 & 250                 & 1000                \\ \hline
None               & Adamatch               & 64.81±1.29          & 67.3±0.53           & 69.42±0.42          & 88.13±2.55          & 85.21±1.44          & 88.33±1.94          & 45.03±4.05          & 52.39±1.65          & 56.18±0.45          \\ \hline
\multirow{4}{*}{A} & Zero-shot              & 66.63±0.00          & 66.63±0.00          & 66.63±0.00          & 91.43±0.00          & 91.43±0.00          & 91.43±0.00          & 52.37±0.00          & 52.37±0.00          & 52.37±0.00          \\
                   & Pseudo-supervise       & 65.79±0.18          & 66.31±0.46          & 65.92±0.33          & 91.37±0.07          & 91.22±0.08          & 91.42±0.08          & 51.11±0.27          & 50.86±0.27          & 50.41±0.41          \\
                   & PL feature input & 67.93±0.80          & 69.6±0.39           & 70.82±0.25          & 91.43±5.51          & \textbf{91.43±0.01} & 90.70±0.85          & 50.00±1.01          & 54.56±0.67          & 56.53±0.61          \\
                   & ZeroMatch   (ours)     & \textbf{69.78±0.94} & \textbf{70.51±0.19} & \textbf{71.81±0.07} & \textbf{91.51±0.12} & 91.42±0.21          & \textbf{91.62±0.15} & \textbf{58.83±0.51} & \textbf{59.69±0.60} & \textbf{60.95±0.13} \\ \hline
\multirow{4}{*}{B} & Zero-shot              & 35.29±0.00          & 35.29±0.00          & 35.29±0.00          & 87.06±0.00          & 87.06±0.00          & 87.06±0.00          & 35.7±0.00           & 35.7±0.00           & 35.7±0.00           \\
                   & Pseudo-supervise       & 36.38±0.51          & 36.29±0.38          & 36.37±0.51          & 88.7±0.16           & 88.51±0.20          & 88.41±0.18          & 36.02±0.04          & 35.97±0.05          & 36.04±0.03          \\
                   & PL feature input & 65.52±0.48          & 65.88±0.89          & 69.29±0.53          & 84.61±2.08          & 86.45±1.55          & 86.66±1.01          & 48.93±1.89          & 53.76±1.62          & 56.06±0.30          \\
                   & ZeroMatch   (ours)     & \textbf{67.05±0.76} & \textbf{67.13±0.73} & \textbf{69.61±0.22} & \textbf{90.42±0.19} & \textbf{90.24±0.08} & \textbf{90.25±0.16} & \textbf{50.23±1.35} & \textbf{53.78±1.74} & \textbf{56.12±0.64} \\ \hline
\end{tabular}
}
\label{tab:nlp_flan_results}
\end{table*}

We also present benchmark results with pseudo-labels from FLAN-T5 and CLIP.
While FLAN-T5 and CLIP models are generally less powerful models compared with GPT-4 and Llama models, this configuration demonstrates following practical settings:
\begin{itemize}
\item \textbf{Unseen task data}: user may train with personal data that is not seen from the foundation models.
Based on training data sources listed in technical reports of FLAN-T5~\cite{chung2024scaling} and CLIP~\cite{radford2021learning} models, we ensure that these models are not trained with our benchmark dataset. On the other hand, recent models like GPT and LLama does not reveal the data source, therfore we cannot ensure that our benchmark datasets (which is public) are not in training the source of these models.
\item \textbf{Restrictions in model access}: user may not be able to access state-of-the-art large foundation models and be limited to using certain types of less powerful foundation models, due to company restrictions, contract, or high implementation cost involved in accomodating the new model.
\end{itemize}
NLP benchmark results with FLAN-T5 are presented in Table~\ref{tab:nlp_flan_results}.
Image classification results with CLIP models are presented in Table~\ref{tab:image_clip_results} in Appendix~\ref{sec:image_classification_with_clip} due to space constraint.
Our results shows that ZeroMatch consitently outperforms or maintains the performance of baseline within one standard deviation all other baselines in most cases (26 of 28 total settings, except CIFAR100 with 400-labels and Flowers102), demonstrating its robustness against low-quaility teacher pseudo-labels.
For example, for Amazon Review with FLAN-T5 pseudo-label set A, ZeroMatch achieves an accuracy of 58.83, outperforming PL feature input's accuracy of 50.00.
Notably, when pseudo-label's quality is low, ZeroMatch consistently improves (or at least maintain the performance within one standard deviation) when number of labels increase.
For example, the scores of Yahoo Answers with FLAN-T5 set B increases from 67.05\% to 69.61\% as number of labels increase.
This implies that our method can utilize labeled samples to correct and improve when pseudo-label may not be helpful.
On the other hand, PL feature input baseline fails to improve in AG News 40-labels and 100-labels setting with FLAN-T5 A set, where the score decreased from 91.43\% to 90.70\%.
Our method remains largely effective against pseudo-supervision under low-quality pseudo-label setting. For example, in 250 label setting of Yahoo Answers, pseudo-supervision reaches $36.38\%$ accuracy with pseudo-label set B, which is worse than $67.05\%$ of ZeroMatch by large margin.

\subsection{A Deeper Look into ZeroMatch}
\label{sec:deeper_look}

\paragraph{Accomodating embeddings can further improve ZeroMatch}
While our problem setup focuses on leveraging pseudo-labels from foundation models, embeddings from foundation models (trained with embedding tasks) may also provide information often helpful for learning the task.
We also provide a way to accomodate these embeddings in our framework to further improve performance on target tasks.
Please see Appendix~\ref{sec:imp_detail_baseline} for implementation details.
We provide benchmarks of our method leveraging pseudo-labels of GPT-4.1 and embeddings from CLIP-large model on Flowers102 and Resisc45 in Table~\ref{tab:add_embedding_flowers}.
Our result shows that adding embedding to our framework can outperform cases where only pseudo-labels from same foundation models are included, achieving highest score in both dataset.
This indicates that our method can be easily extended to mix and match pseudo-labels with embeddings to further improve the performance.

\begin{table*}[t!]
    \centering
    \small
    \caption{Accuracy (\%) in Flowers102, Resisc45 with pseudo-labels and embeddings from CLIP models and GPT-4.1. Median and standard deviation of 3 different random seeds are reported.}
    \begin{tabular}{c|c|c|c}
    \hline
    Method           & FM inference info used                                                                 & Flowers102          & Resisc45              \\ \hline
    Zero-shot        & GPT-4.1 pseudo-label                                                                   & 88.37±0.00          & 79.28±0.00          \\
    Zero-shot        & CLIP-large pseudo-label                                                                & 72.13±0.00          & 60.32±0.00          \\
    ZeroMatch (ours) & GPT-4.1 pseudo-label                                                                   & 95.17±0.88          & 87.65±0.73          \\ \hline
    ZeroMatch (ours) & \begin{tabular}[c]{@{}c@{}}GPT-4.1 pseudo-label +\\  CLIP-large embedding\end{tabular} & \textbf{97.20±0.41} & \textbf{88.52±0.77} \\ \hline
    \end{tabular}
    \label{tab:add_embedding_flowers}
\end{table*}

\paragraph{Ablation study}

Two main components that constructs our method (other than SSL training) is separate knowledge disillation (KD) stage, and learning KD as auxiliary loss for SSL's objective. Effectiveness of each component is validated through following ablation study presented in Appendix~\ref{sec:ablation_study}.
Our results shows that adding auxiliary KD loss helps, especially in the low-label setting, and running ZeroMatch without stage 1 can sometimes achieve the similar performance to original ZeroMatch, but it may also produce large gap in performance, depending on the dataset.

\paragraph{Hyperparameter sensitivity}
\label{sec:hyperparameter_sensitivity}
While we mainly benchmark our method with hyperparameters $\alpha_p=1$ and $\lambda_p=1$ without finding the optimal setting,
we also provide a sensitivity analysis on combinations of these parameters in Appendix~\ref{sec:hyperparameter_sensitivity}.
We find that while $(\alpha_p, \lambda_p) = (1, 1.0)$ can be a good candidate on some datasets (ex. Yahoo Answers), there may exist other optimal parameters that can further improve performance.

\paragraph{Computational cost analysis}
\label{sec:compute_cost}
We discuss comparison of memory usage and computation cost associated with our method and SSL baseline(AdaMatch) in Appendix~\ref{sec:compute_cost}.
Our method consumes around 50\% more training time compared with AdaMatch due to KD stage, and uses slightly more VRAM than AdaMatch.
Nevertheless, we note that VRAM usage of ZeroMatch is less than 8GB, making it suitable to run in low compute devices.
We also provide possible alternative implementation of ZeroMatch to reduce the compute cost.


\section{Conclusion}

This work introduces ZeroMatch, a framework for robustly integrating foundation model predictions into semi-supervised learning. Through extensive experiments across multiple datasets from different domains, we demonstrate that ZeroMatch achieves state-of-the-art performance while maintaining robustness to varying pseudo-label quality. Our learning-based mechanism effectively balances between limited labels and plentiful pseudo-labels, enabling practitioners to leverage foundation models without the computational overhead of direct fine-tuning. Future work could explore extending ZeroMatch to more complex tasks and investigating theoretical guarantees for the weighting mechanism.

\clearpage
\bibliographystyle{unsrt}
\bibliography{ref}

\begin{thebibliography}{10}

\bibitem{northcutt1pervasive}
Curtis~G Northcutt, Anish Athalye, and Jonas Mueller.
\newblock Pervasive label errors in test sets destabilize machine learning
  benchmarks.
\newblock In {\em Thirty-fifth Conference on Neural Information Processing
  Systems Datasets and Benchmarks Track (Round 1)}, 2021.

\bibitem{sun2017revisiting}
Chen Sun, Abhinav Shrivastava, Saurabh Singh, and Abhinav Gupta.
\newblock Revisiting unreasonable effectiveness of data in deep learning era.
\newblock In {\em Proceedings of the IEEE international conference on computer
  vision}, pages 843--852, 2017.

\bibitem{shen2024data}
Judy~Hanwen Shen, Inioluwa~Deborah Raji, and Irene~Y Chen.
\newblock The data addition dilemma.
\newblock In {\em Machine Learning for Healthcare Conference}. PMLR, 2024.

\bibitem{tarvainen2017mean}
Antti Tarvainen and Harri Valpola.
\newblock Mean teachers are better role models: Weight-averaged consistency
  targets improve semi-supervised deep learning results.
\newblock {\em Advances in neural information processing systems}, 30, 2017.

\bibitem{sohn2020fixmatch}
Kihyuk Sohn, David Berthelot, Nicholas Carlini, Zizhao Zhang, Han Zhang,
  Colin~A Raffel, Ekin~Dogus Cubuk, Alexey Kurakin, and Chun-Liang Li.
\newblock Fixmatch: Simplifying semi-supervised learning with consistency and
  confidence.
\newblock {\em Advances in neural information processing systems}, 33:596--608,
  2020.

\bibitem{laine2022temporal}
Samuli Laine and Timo Aila.
\newblock Temporal ensembling for semi-supervised learning.
\newblock In {\em International Conference on Learning Representations}, 2022.

\bibitem{brown2020language}
Tom Brown, Benjamin Mann, Nick Ryder, Melanie Subbiah, Jared~D Kaplan, Prafulla
  Dhariwal, Arvind Neelakantan, Pranav Shyam, Girish Sastry, Amanda Askell,
  et~al.
\newblock Language models are few-shot learners.
\newblock {\em Advances in neural information processing systems},
  33:1877--1901, 2020.

\bibitem{liangholistic}
Percy Liang, Rishi Bommasani, Tony Lee, Dimitris Tsipras, Dilara Soylu,
  Michihiro Yasunaga, Yian Zhang, Deepak Narayanan, Yuhuai Wu, Ananya Kumar,
  et~al.
\newblock Holistic evaluation of language models.
\newblock {\em Transactions on Machine Learning Research}, 2022.

\bibitem{shi2023rethinking}
Zhengxiang Shi, Francesco Tonolini, Nikolaos Aletras, Emine Yilmaz, Gabriella
  Kazai, and Yunlong Jiao.
\newblock Rethinking semi-supervised learning with language models.
\newblock In {\em Findings of the Association for Computational Linguistics:
  ACL 2023}, pages 5614--5634, 2023.

\bibitem{zhang2024candidate}
Jiahan Zhang, Qi~Wei, Feng Liu, and Lei Feng.
\newblock Candidate pseudolabel learning: Enhancing vision-language models by
  prompt tuning with unlabeled data.
\newblock {\em arXiv preprint arXiv:2406.10502}, 2024.

\bibitem{ganerasing}
Kai Gan and Tong Wei.
\newblock Erasing the bias: Fine-tuning foundation models for semi-supervised
  learning.
\newblock In {\em Forty-first International Conference on Machine Learning},
  2024.

\bibitem{hegselmann2023tabllm}
Stefan Hegselmann, Alejandro Buendia, Hunter Lang, Monica Agrawal, Xiaoyi
  Jiang, and David Sontag.
\newblock Tabllm: Few-shot classification of tabular data with large language
  models.
\newblock In {\em International Conference on Artificial Intelligence and
  Statistics}, pages 5549--5581. PMLR, 2023.

\bibitem{nam2023semi}
Jaehyun Nam, Woomin Song, Seong~Hyeon Park, Jihoon Tack, Sukmin Yun, Jaehyung
  Kim, and Jinwoo Shin.
\newblock Semi-supervised tabular classification via in-context learning of
  large language models.
\newblock In {\em Workshop on Efficient Systems for Foundation Models@
  ICML2023}, 2023.

\bibitem{yang2025knowledge}
Jing Yang, Xiatian Zhu, Adrian Bulat, Brais Martinez, and Georgios
  Tzimiropoulos.
\newblock Knowledge distillation meets open-set semi-supervised learning.
\newblock {\em International Journal of Computer Vision}, 133(1):315--334,
  2025.

\bibitem{shiselkd}
Liangliang Shi, Zhengyan Shi, and Junchi Yan.
\newblock Selkd: Selective knowledge distillation via optimal transport
  perspective.
\newblock In {\em The Thirteenth International Conference on Learning
  Representations}.

\bibitem{vemulapalli2023knowledge}
Raviteja Vemulapalli, Hadi Pouransari, Fartash Faghri, Sachin Mehta, Mehrdad
  Farajtabar, Mohammad Rastegari, and Oncel Tuzel.
\newblock Knowledge transfer from vision foundation models for efficient
  training of small task-specific models.
\newblock {\em arXiv preprint arXiv:2311.18237}, 2023.

\bibitem{zhao2023multistage}
Jiachen Zhao, Wenlong Zhao, Andrew Drozdov, Benjamin Rozonoyer, Md~Arafat
  Sultan, Jay-Yoon Lee, Mohit Iyyer, and Andrew McCallum.
\newblock Multistage collaborative knowledge distillation from a large language
  model for semi-supervised sequence generation.
\newblock {\em arXiv preprint arXiv:2311.08640}, 2023.

\bibitem{jiang2023disco}
Weifeng Jiang, Qianren Mao, Chenghua Lin, Jianxin Li, Ting Deng, Weiyi Yang,
  and Zheng Wang.
\newblock Disco: distilled student models co-training for semi-supervised text
  mining.
\newblock {\em arXiv preprint arXiv:2305.12074}, 2023.

\bibitem{zhu2024doubly}
Banghua Zhu, Mingyu Ding, Philip Jacobson, Ming Wu, Wei Zhan, Michael Jordan,
  and Jiantao Jiao.
\newblock Doubly-robust self-training.
\newblock {\em Advances in Neural Information Processing Systems}, 36, 2024.

\bibitem{rizvedefense}
Mamshad~Nayeem Rizve, Kevin Duarte, Yogesh~S Rawat, and Mubarak Shah.
\newblock In defense of pseudo-labeling: An uncertainty-aware pseudo-label
  selection framework for semi-supervised learning.
\newblock In {\em International Conference on Learning Representations}, 2021.

\bibitem{zhang2021flexmatch}
Bowen Zhang, Yidong Wang, Wenxin Hou, Hao Wu, Jindong Wang, Manabu Okumura, and
  Takahiro Shinozaki.
\newblock Flexmatch: Boosting semi-supervised learning with curriculum pseudo
  labeling.
\newblock {\em Advances in Neural Information Processing Systems},
  34:18408--18419, 2021.

\bibitem{zheng2022simmatch}
Mingkai Zheng, Shan You, Lang Huang, Fei Wang, Chen Qian, and Chang Xu.
\newblock Simmatch: Semi-supervised learning with similarity matching.
\newblock In {\em Proceedings of the IEEE/CVF Conference on Computer Vision and
  Pattern Recognition}, pages 14471--14481, 2022.

\bibitem{you2024diffusion}
Zebin You, Yong Zhong, Fan Bao, Jiacheng Sun, Chongxuan Li, and Jun Zhu.
\newblock Diffusion models and semi-supervised learners benefit mutually with
  few labels.
\newblock {\em Advances in Neural Information Processing Systems}, 36, 2024.

\bibitem{menghini2023enhancing}
Cristina Menghini, Andrew Delworth, and Stephen Bach.
\newblock Enhancing clip with clip: Exploring pseudolabeling for limited-label
  prompt tuning.
\newblock {\em Advances in Neural Information Processing Systems},
  36:60984--61007, 2023.

\bibitem{buciluǎ2006model}
Cristian Buciluǎ, Rich Caruana, and Alexandru Niculescu-Mizil.
\newblock Model compre ssion.
\newblock In {\em Proceedings of the 12th ACM SIGKDD international conference
  on Knowledge discovery and data mining}, pages 535--541, 2006.

\bibitem{hinton2015distilling}
Geoffrey Hinton, Oriol Vinyals, and Jeff Dean.
\newblock Distilling the knowledge in a neural network.
\newblock {\em arXiv preprint arXiv:1503.02531}, 2015.

\bibitem{west2021symbolic}
Peter West, Chandra Bhagavatula, Jack Hessel, Jena~D Hwang, Liwei Jiang,
  Ronan~Le Bras, Ximing Lu, Sean Welleck, and Yejin Choi.
\newblock Symbolic knowledge distillation: from general language models to
  commonsense models.
\newblock {\em arXiv preprint arXiv:2110.07178}, 2021.

\bibitem{beyer2022knowledge}
Lucas Beyer, Xiaohua Zhai, Am{\'e}lie Royer, Larisa Markeeva, Rohan Anil, and
  Alexander Kolesnikov.
\newblock Knowledge distillation: A good teacher is patient and consistent.
\newblock In {\em Proceedings of the IEEE/CVF conference on computer vision and
  pattern recognition}, pages 10925--10934, 2022.

\bibitem{hsieh2023distilling}
Cheng-Yu Hsieh, Chun-Liang Li, Chih-Kuan Yeh, Hootan Nakhost, Yasuhisa Fujii,
  Alexander Ratner, Ranjay Krishna, Chen-Yu Lee, and Tomas Pfister.
\newblock Distilling step-by-step! outperforming larger language models with
  less training data and smaller model sizes.
\newblock {\em arXiv preprint arXiv:2305.02301}, 2023.

\bibitem{voge2024leveraging}
Lukas V{\"o}ge, Vincent Gurgul, and Stefan Lessmann.
\newblock Leveraging zero-shot prompting for efficient language model
  distillation.
\newblock {\em arXiv preprint arXiv:2403.15886}, 2024.

\bibitem{chen2020big}
Ting Chen, Simon Kornblith, Kevin Swersky, Mohammad Norouzi, and Geoffrey~E
  Hinton.
\newblock Big self-supervised models are strong semi-supervised learners.
\newblock {\em Advances in neural information processing systems},
  33:22243--22255, 2020.

\bibitem{he2021semi}
Lingxiao He, Wu~Liu, Jian Liang, Kecheng Zheng, Xingyu Liao, Peng Cheng, and
  Tao Mei.
\newblock Semi-supervised domain generalizable person re-identification.
\newblock {\em arXiv preprint arXiv:2108.05045}, 2021.

\bibitem{xie2022spd}
Bangquan Xie, Zongming Yang, Liang Yang, Ruifa Luo, Jun Lu, Ailin Wei,
  Xiaoxiong Weng, and Bing Li.
\newblock Spd: Semi-supervised learning and progressive distillation for 3-d
  detection.
\newblock {\em IEEE Transactions on Neural Networks and Learning Systems},
  35(3):3503--3513, 2022.

\bibitem{lee2013pseudolabel}
Dong-Hyun Lee.
\newblock Pseudo-label : The simple and efficient semi-supervised learning
  method for deep neural networks.
\newblock 2013.

\bibitem{qizhe2020}
Qizhe Xie, Zihang Dai, Eduard Hovy, Thang Luong, and Quoc Le.
\newblock Unsupervised data augmentation for consistency training.
\newblock In H.~Larochelle, M.~Ranzato, R.~Hadsell, M.F. Balcan, and H.~Lin,
  editors, {\em Advances in Neural Information Processing Systems}, volume~33,
  pages 6256--6268. Curran Associates, Inc., 2020.

\bibitem{berthelot2020}
David Berthelot, Nicholas Carlini, Ekin~D. Cubuk, Alex Kurakin, Kihyuk Sohn,
  Han Zhang, and Colin Raffel.
\newblock Remixmatch: Semi-supervised learning with distribution matching and
  augmentation anchoring.
\newblock In {\em 8th International Conference on Learning Representations,
  {ICLR} 2020, Addis Ababa, Ethiopia, April 26-30, 2020}. OpenReview.net, 2020.

\bibitem{goodfellow2013empirical}
Ian~J Goodfellow, Mehdi Mirza, Da~Xiao, Aaron Courville, and Yoshua Bengio.
\newblock An empirical investigation of catastrophic forgetting in
  gradient-based neural networks.
\newblock {\em arXiv preprint arXiv:1312.6211}, 2013.

\bibitem{kar2022preventing}
Sudipta Kar, Giuseppe Castellucci, Simone Filice, Shervin Malmasi, and Oleg
  Rokhlenko.
\newblock Preventing catastrophic forgetting in continual learning of new
  natural language tasks.
\newblock In {\em Proceedings of the 28th ACM SIGKDD Conference on Knowledge
  Discovery and Data Mining}, pages 3137--3145, 2022.

\bibitem{berthelotadamatch}
David Berthelot, Rebecca Roelofs, Kihyuk Sohn, Nicholas Carlini, and Alexey
  Kurakin.
\newblock Adamatch: A unified approach to semi-supervised learning and domain
  adaptation.
\newblock In {\em International Conference on Learning Representations}, 2021.

\bibitem{wang2022usb}
Yidong Wang, Hao Chen, Yue Fan, Wang Sun, Ran Tao, Wenxin Hou, Renjie Wang,
  Linyi Yang, Zhi Zhou, Lan-Zhe Guo, et~al.
\newblock Usb: A unified semi-supervised learning benchmark for classification.
\newblock {\em Advances in Neural Information Processing Systems},
  35:3938--3961, 2022.

\bibitem{openai2024gpt4o}
OpenAI.
\newblock Gpt-4o technical report, 2024.
\newblock Accessed: 2025-05-15.

\bibitem{meta2024llama33}
Meta AI.
\newblock Llama 3.3-70b instruct model card, 2024.
\newblock Accessed: 2025-05-15.

\bibitem{chung2024scaling}
Hyung~Won Chung, Le~Hou, Shayne Longpre, Barret Zoph, Yi~Tay, William Fedus,
  Yunxuan Li, Xuezhi Wang, Mostafa Dehghani, Siddhartha Brahma, et~al.
\newblock Scaling instruction-finetuned language models.
\newblock {\em Journal of Machine Learning Research}, 25(70):1--53, 2024.

\bibitem{openai2025gpt41}
OpenAI.
\newblock Introducing gpt-4.1 in the api.
\newblock \url{https://openai.com/index/gpt-4-1/}, 2025.
\newblock Accessed: 2025-05-15.

\bibitem{radford2021learning}
Alec Radford, Jong~Wook Kim, Chris Hallacy, Aditya Ramesh, Gabriel Goh,
  Sandhini Agarwal, Girish Sastry, Amanda Askell, Pamela Mishkin, Jack Clark,
  et~al.
\newblock Learning transferable visual models from natural language
  supervision.
\newblock In {\em International conference on machine learning}, pages
  8748--8763. PMLR, 2021.

\bibitem{dosovitskiy2020image}
Alexey Dosovitskiy.
\newblock An image is worth 16x16 words: Transformers for image recognition at
  scale.
\newblock {\em arXiv preprint arXiv:2010.11929}, 2020.

\bibitem{devlin2018bert}
Jacob Devlin.
\newblock Bert: Pre-training of deep bidirectional transformers for language
  understanding.
\newblock {\em arXiv preprint arXiv:1810.04805}, 2018.

\bibitem{wang2022generalizing}
Jindong Wang, Cuiling Lan, Chang Liu, Yidong Ouyang, Tao Qin, Wang Lu, Yiqiang
  Chen, Wenjun Zeng, and S~Yu Philip.
\newblock Generalizing to unseen domains: A survey on domain generalization.
\newblock {\em IEEE transactions on knowledge and data engineering},
  35(8):8052--8072, 2022.

\bibitem{borisov2022deep}
Vadim Borisov, Tobias Leemann, Kathrin Se{\ss}ler, Johannes Haug, Martin
  Pawelczyk, and Gjergji Kasneci.
\newblock Deep neural networks and tabular data: A survey.
\newblock {\em IEEE transactions on neural networks and learning systems},
  2022.

\bibitem{hu2022lora}
Edward~J Hu, Yelong Shen, Phillip Wallis, Zeyuan Allen-Zhu, Yuanzhi Li, Shean
  Wang, Lu~Wang, Weizhu Chen, et~al.
\newblock Lora: Low-rank adaptation of large language models.
\newblock {\em ICLR}, 1(2):3, 2022.

\bibitem{lester2021power}
Brian Lester, Rami Al-Rfou, and Noah Constant.
\newblock The power of scale for parameter-efficient prompt tuning.
\newblock {\em arXiv preprint arXiv:2104.08691}, 2021.

\bibitem{aiindex2025}
AI~Index~Steering Committee.
\newblock Ai index report 2025.
\newblock \url{https://hai.stanford.edu/ai-index/2025-ai-index-report}, 2025.
\newblock \url{https://hai.stanford.edu/ai-index/2025-ai-index-report}.

\bibitem{Krizhevsky09learningmultiple}
Alex Krizhevsky.
\newblock Learning multiple layers of features from tiny images.
\newblock Technical report, 2009.

\bibitem{nilsback2008automated}
Maria-Elena Nilsback and Andrew Zisserman.
\newblock Automated flower classification over a large number of classes.
\newblock In {\em 2008 Sixth Indian conference on computer vision, graphics \&
  image processing}, pages 722--729. IEEE, 2008.

\bibitem{cheng2017remote}
Gong Cheng, Junwei Han, and Xiaoqiang Lu.
\newblock Remote sensing image scene classification: Benchmark and state of the
  art.
\newblock {\em Proceedings of the IEEE}, 105(10):1865--1883, 2017.

\bibitem{chang2008importance}
Ming-Wei Chang, Lev Ratinov, Dan Roth, and Vivek Srikumar.
\newblock Importance of semantic representation: dataless classification.
\newblock In {\em Proceedings of the 23rd National Conference on Artificial
  Intelligence - Volume 2}, AAAI'08, page 830–835. AAAI Press, 2008.

\bibitem{zhang2015character}
Xiang Zhang, Junbo Zhao, and Yann LeCun.
\newblock Character-level convolutional networks for text classification.
\newblock {\em Advances in neural information processing systems}, 28, 2015.

\bibitem{wang2024qwen2}
Peng Wang, Shuai Bai, Sinan Tan, Shijie Wang, Zhihao Fan, Jinze Bai, Keqin
  Chen, Xuejing Liu, Jialin Wang, Wenbin Ge, et~al.
\newblock Qwen2-vl: Enhancing vision-language model's perception of the world
  at any resolution.
\newblock {\em arXiv preprint arXiv:2409.12191}, 2024.

\bibitem{cubuk2020randaugment}
Ekin~D Cubuk, Barret Zoph, Jonathon Shlens, and Quoc~V Le.
\newblock Randaugment: Practical automated data augmentation with a reduced
  search space.
\newblock In {\em Proceedings of the IEEE/CVF conference on computer vision and
  pattern recognition workshops}, pages 702--703, 2020.

\bibitem{zhuincorporating}
Jinhua Zhu, Yingce Xia, Lijun Wu, Di~He, Tao Qin, Wengang Zhou, Houqiang Li,
  and Tieyan Liu.
\newblock Incorporating bert into neural machine translation.
\newblock In {\em International Conference on Learning Representations}, 2020.

\bibitem{jia2022visual}
Menglin Jia, Luming Tang, Bor-Chun Chen, Claire Cardie, Serge Belongie, Bharath
  Hariharan, and Ser-Nam Lim.
\newblock Visual prompt tuning.
\newblock In {\em European conference on computer vision}, pages 709--727.
  Springer, 2022.

\bibitem{zhou2022learning}
Kaiyang Zhou, Jingkang Yang, Chen~Change Loy, and Ziwei Liu.
\newblock Learning to prompt for vision-language models.
\newblock {\em International Journal of Computer Vision}, 130(9):2337--2348,
  2022.

\end{thebibliography}


\appendix
\onecolumn

\section{Additional details for: When is Our Setup Most Useful?}
\label{sec:setup_useful}

In this section, we provide further discussion on where our setup becomes particularly useful.

\paragraph{Our Setup can Leverage Large FMs for SSL with Limited Compute}
In terms of the statistical performance, fine-tuning (or parameter-efficient fine-tuning~\cite{hu2022lora,lester2021power}) open-weight large FM is often the best approach, as it allows models to fully exploit their rich parameterization. However, this approach is resource-intensive due to large parameter size and costly backpropagation of LFMs—for example, fine-tuning LLaMA-70B with LoRA and 16-bit quantization requires over 160GB of GPU memory, making it impractical for users without access to high-end GPU clusters. In contrast, our setup enables the use of LFM in SSL fine-tuning task of smaller models on modest hardware, such as a single GPU or even edge devices. This decouples the use of high quality information of LFMs from the computational demands of training, allowing effective model training locally without incurring high computational costs.

\paragraph{Our Setup is Compatible with Closed-weight FMs}
Many high-performing FMs (e.g., GPT-4o) are closed-weight and accessible only through limited APIs. These APIs often restrict training capabilities—for instance, OpenAI's fine-tuning API supports only supervised learning with fixed input-output pairs, making training with SSL algorithm infeasible.
Our setup remains compatible in such scenarios. Extracting pseudo-labels from the FM’s text output is straightforward and does not require access to internal representations or gradients. This makes our approach practical even when constrained to inference-only interactions with proprietary FMs.

\paragraph{Our Setup is Reduces the Risk of Data Leakage}
Using LFMs often requires uploading data to remote servers where the model is hosted, raising concerns of data leak—especially when working with sensitive or proprietary datasets. Assuming that the model training happens in private device, our setup significantly reduces this risk by limiting data exposure to remote machines to one-time inference call. Once the data is processed, it can be immediately discarded, eliminating the need for persistent storage or repeated transmission.
This is notably safer than fine-tuning LFMs, which requires continuous access to the training data throughout its training process.

\subsection{Using Pseudo-labels vs Embedding}
\label{sub:pseudo_labels_vs_embedding}
Among the available outputs of foundation models that can be obtained from inference (e.g., embeddings, logits, reasonings), our problem setup focus on zero-shot prompted pseudo-labels for their high performance, simplicity, and human interpretability.
Zero-shot classification is one of the most fastest growing application with rising performance, and most frequently updated to accomodate recent knowledge~\cite{aiindex2025}. Additionally, the pseudo-label outputs are inherently interpretable by humans, making them favorable for use in downstream tasks and evaluations.
In contrast, leveraging higher-dimensional outputs like embeddings, logits, or reasonings may provide richer information in some tasks,  these representations often capture more information than necessary for the task at hand, introducing additional complexity when used as supervision targets in knowledge distillation or as inputs for downstream models.
Due to these reasons, we prioritize zero-shot pseudo-labels in this work.
Additionally, we briefly discuss methods to involve embeddings in our framework in Section~\ref{sec:deeper_look}.
We note that in-context learning with few labeled examples could also be used to generate pseudo-labels for our setting.

\section{Experiment Details}
\label{sec:appendix-experiment-detail}

\subsection{Datasets}
\label{sec:appendix-data}

We use six datasets from NLP and Image classification domain. We provide description of each of them.

\paragraph{CIFAR-100}
CIFAR-100~\cite{Krizhevsky09learningmultiple} dataset is a (32×32 pixels) natural image recognition dataset consisting 100 classes. There are 500 training samples and 100 test samples per class.

\paragraph{Flowers102}
Flowers102~\cite{nilsback2008automated} is a image dataset containing 102 flower categories commonly found in the United Kingdom.
The dataset contains minumum 40 to maximum 258 images per class.

\paragraph{Resisc45}
RESICS45~\cite{cheng2017remote} is a publicly available benchmark for Remote Sensing Image Scene Classification. Images from 45 kind of scenes are presented.

\paragraph{Yahoo! Answer}
Yahoo! Answer~\cite{chang2008importance} is a question text topic classification dataset with 10 categories of topics.
Each class contains 140,000 training samples and 6,000 test samples.
Training and validation set is drawn by sampling 50,000 samples and 5,000 samples per class from training samples, and test set is unchanged, following the data settings in USB~\cite{wang2022generalizing}.

\paragraph{AG News}
The AG News~\cite{zhang2015character} dataset is a news text topic classification dataset with 4 classes.
Original dataset contains 30,000 training samples and 1,900 test samples per class.
In our experiment 25,000 samples and 2,500 samples per class are sampled from training samples following USB~\cite{wang2022generalizing}. The test dataset is unchanged.

\paragraph{Amazon Review}
The Amazon Review [52] dataset is a sentiment classification dataset with product review text input.
There are 5 classes indicating the review score. Each class (score) contains 600,000 training samples and 130,000 test samples. Following USB~\cite{wang2022generalizing}, 50,000 samples and 5,000 samples per class from training samples are used as training dataset and validation dataset respectively. The test dataset is unchanged.

We use the data splits provided by USB~\cite{wang2022usb} for NLP datasets, and splits provided by CPL~\cite{zhang2024candidate} for Flowers102 and Resisc45.

\subsection{Obtaining Pseudo-labels from Foundation Models}
\label{sec:plabel}

In this section, we discuss details in obtaining the zero-shot pseudo-labels to train with ZeroMatch.

\subsubsection{Details of pseudo-label generation process}

\paragraph{Image classification datasets}
For image classification datasets, we mainly use GPT-4.1 and CLIP models to generate pseudo-labels.

For GPT-4.1~\cite{openai2025gpt41}, we use chat completion feature to prompt for image classification.
We use following system prompt to describe for the task we need:
\begin{table}[h]
\centering
\begin{tabular}{|l|}
\hline
\begin{tabular}[c]{@{}l@{}}You are an image classification system, asked to classify \textit{\{topic\}}.\\
Classify the input image into exactly one of the following types: \textit{\{candidate labels\}} \\
Respond with only the type that best fits the image. \\
\end{tabular} \\ \hline
\end{tabular}
\end{table}

Where the topic variable is set as `flowers' for Flowers102, `scene' for Resisc45, and `tiny image' for CIFAR100. Candidate labels are set as possible labels in text format.
Same procedure is applied when using other chat-completion based models (GPT-4o~\cite{openai2024gpt4o}, Qwen2-VL~\cite{wang2024qwen2}, Llama3.2-Vision~\cite{meta2024llama33}).
After obtaining the output, we parse the output text to see if text labels in candidate label set exists in output text. In cases where no label is detected, we use default pseudo-label: \texttt{rose} for Flowers102, \texttt{commercial area} for Resisc45, and \texttt{road} for CIFAR100.

When using CLIP model, we follow the procedures of zero-shot classification in ~\cite{radford2021learning}.
We compute the feature embedding of a given image, and for each label in candidate labels, we prompt the model with following prompt: \texttt{This is a photo of \{label\}} and get feature embedding.
We then obtain the pseudo-label by finding the label that gives closest embedding to the image embedding, based on cosine similarity.

\paragraph{NLP classification datasets - Topic classification: AG News, Yahoo Answers}
To obtain pseudo-labels for topic classification tasks, we use following prompting template with GPT-4o, Llama3.3-70B, and FLAN-T5 models for given input text:

\begin{table}[h]
\centering
\begin{tabular}{|l|}
\hline
\begin{tabular}[c]{@{}l@{}}Select the topic that the given article is about. The topics are: \textit{\{candidate\_labels\}}.\\ \\ Article: \textit{\{input\_text\}},\\ Answer:\end{tabular} \\ \hline
\end{tabular}
\end{table}

For AG News, the candidate labels are \texttt{world, sports, business, technology}, and for Yahoo Answers, the candidate labels are \texttt{society, science, health, education, computer, sports, business, entertainment, relationship, politics}.
In case the output does not contain labels from candidate labels or is not parsable, we set the pseudo-label to be
\texttt{world} for AG News, and \texttt{society} for the Yahoo Answers as the default label.
For GPT-4o and Llama models, we use chat completion template with following system prompt: \texttt{You are a helpful assistant.}

\paragraph{NLP classification datasets - Sentiment classification: Amazon Review}
We use following prompt template with GPT-4o, LLama3.3-70B, and FLAN-T5 models models for given input text to obtain the review text's sentiment:

\begin{table}[h]
\centering
\begin{tabular}{|l|}
\hline
\begin{tabular}[c]{@{}l@{}}\textit{\{input\_text\}}\\ What is the sentiment of this review?\\ \\ OPTIONS:\\ - very negative\\ - negative\\ - neutral\\ - positive\\ - very positive\end{tabular} \\ \hline
\end{tabular}
\end{table}

In case of model output not being recognized as one of candidate labels, we select \texttt{neutral} as the default pseudo-label.
For GPT-4o and LLama models, we use chat completion template with following system prompt: \texttt{You are a helpful assistant.}

\subsubsection{Zero-shot benchmark results and foundation model selection}
\label{sec:zero_shot_benchmark}

To select foundation models to be used with ZeroMatch, we consider 1) best performing large foundation models that can provide maximum statistical performance with our method and 2) lowest quality foundation models to test robustness of our method against inaccurate pseudo-labels.

We provide zero-shot benchmark results on test set in Table~\ref{tab:zero_shot_img} (NLP) and Table~\ref{tab:zero_shot_nlp} (Image).

\begin{table*}[ht]
    \centering
    \small
    \caption{
    Zero-shot pseudo-label accuracies(\%) for NLP datasets (test set).
    }
    \begin{tabular}{c|ccc}
    \hline
    Dataset       & Yahoo Answers  & AG News        & Amazon Review  \\ \hline
    GPT-4o        & 68.81          & 86.25          & \textbf{59.14} \\
    Llama3.3-70B  & \textbf{69.15} & \textbf{88.41} & 55.79          \\ \hline
    FLAN-T5-XXL   & \textbf{66.62}          & \textbf{91.43}          & 45.61          \\
    FLAN-T5-XL    & 63.97          & 91.39          & \textbf{52.37}          \\
    FLAN-T5-large & 61.33          & 87.66          & 51.96          \\
    FLAN-T5-base  & 55.17          & 88.68          & 42.23          \\
    FLAN-T5-small & 29.44          & 87.07          & 35.7           \\ \hline
    \end{tabular}
    \label{tab:zero_shot_img}
\end{table*}

We found that GPT-4o and LLama3.3-70B were highest performing overall, scoring top-2 on all the datasets.
Among FLAN-T5 models, we find that FLAN-T5-XXL performs the best in Yahoo Answers and AG News dataset, while FLAN-T5-XL model performed better in Amazon review.
To benchmark our method with highest and lowest accuracy, FLAN-T5-XXL and FLAN-T5-small were selected as A and B set of pseudo-labels in AG News and Yahoo Answers, and for Amazon reivew, we use FLAN-T5-XL and FLAN-T5-small as A and B set.

\begin{table*}[ht]
    \centering
    \small
    \caption{
    Zero-shot pseudo-label accuracies(\%) for image datasets (test set).
    }
    \begin{tabular}{c|ccc}
    \hline
    Dataset             & CIFAR100       & Flowers102     & Resisc45       \\ \hline
    GPT-4.1             & \textbf{83.25} & \textbf{88.37} & \textbf{79.28} \\
    GPT-4o              & 81.5           & 86.0            & 73.46          \\
    Qwen2-VL            & 62.58          & 78.11          & 68.25          \\
    Llama3.2-90B-Vision & 36.05          & 59.02          & 44.58          \\ \hline
    CLIP-large-patch14  & 62.27          & 72.13          & 60.32          \\
    CLIP-base-patch32   & 49.49          & 60.17          & 49.63          \\ \hline
    \end{tabular}
    \label{tab:zero_shot_nlp}
\end{table*}

For large foundation models, we benchmarked GPT-4.1, GPT-4o, Qwen2-VL, Llama3.2-90B-Vision with our target datasets, and found that GPT-4.1 was best in all the tasks. As a result we accomodate GPT-4.1, CLIP-large and CLIP-base models.

\subsection{Training setup}
\label{sec:training-setup}

\paragraph{Setup for Image Classification}
We use ViT~\cite{dosovitskiy2020image} as the backbone of our model and employ pre-trained ViT-Small models, provided by the USB framework~\cite{wang2022generalizing}.
For CIFAR100, ViT-Small with patch size of 2 with image size of 32 is used, and for Flowers102 and Resisc45 datasets, ViT-Small with patch size of 16 with image size of 224 is used.
The optimization is performed using the AdamW optimizer with a cosine learning rate schedule given by $\eta_t = \eta_0 \cos \left( \frac{7\pi t}{16T}\right)$ where the initial learning rate is set to $\eta_0 = 5\times10^{-4}$ for CIFAR100 and $\eta_0 = 1\times10^{-4}$ for Flowers102 and Resisc45.
For CIFAR100, we train the model for a total of $T = 204,800$ steps, with a warm-up phase of 5,120 steps, and for Flowers102 and Resisc45, we run for half the size of training steps $T = 102,400$ with 2,560 warm up steps to reflect the smaller training set size.
The batch sizes for both labeled and unlabeled data are set to 16.
For AdaMatch-specific parameters, the threshold for the utilization cutoff mask is set to 0.95.
For data augmentation, we apply random cropping and random horizontal flipping for weak augmentation, while RandAugment~\cite{cubuk2020randaugment} is utilized for strong augmentation.
For knowledge distillation stage (stage 1 of Zeromatch), we run supervised training with teacher pseudo-labels, for same number of training steps.
Regarding the hyperparameters specific to our method, we report results with $\lambda_p = 1$ and with annealing ($\alpha_p=1$). Detailed hyperparameter settings are provided in Table~\ref{tab:hyperparams_img}.

\paragraph{Setup for Natural Language Classification}
For NLP tasks, we fine-tune the pre-trained BERT-Base~\cite{devlin2018bert} using the AdamW optimizer with a cosine learning rate schedule.
The total number of training steps is set to 102,400, with a warm-up phase of 5,120 steps.
Both the labeled and unlabeled batch sizes are set to 4.
All input text is truncated to ensure its length remains within the context length of BERT-Base.
For data augmentation, we employ back-translation~\cite{zhuincorporating} using German-English and Russian-English translation as strong augmentations. No weak augmentation is applied, and the original text input is used instead.
We run AdaMatch with a cutoff threshold of $\tau=0.95$.
Dataset-specific hyperparameters are detailed in Table~\ref{tab:hyperparams_nlp}.
For knowledge distillation stage (stage 1 of Zeromatch), we run supervised training with teacher pseudo-labels, for same number of training steps.
We report the performance of ZeroMatch with $\lambda_p = 1$ and with annealing ($\alpha_p=1$).

\begin{table*}[h]
\centering
\caption{Hyperparameters of Image classification datasets.}
\resizebox{\textwidth}{!}{
\begin{tabular}{c|ccc}
Dataset                 & CIFAR100                                                                       & Flowers102                                                    & Resisc45                                                     \\ \hline
Image Size              & 32                                                                             & 224                                                           & 224                                                          \\
Model                   & ViT-S-P4-32                                                                    & ViT-S-P16-224                                                 & ViT-S-P16-224                                                \\
Weight Decay            & \multicolumn{3}{c}{5e-4}                                                                                                                                                                                      \\
Labeled Batch size      & \multicolumn{3}{c}{16}                                                                                                                                                                                        \\
Unlabeled Batch size    & \multicolumn{3}{c}{16}                                                                                                                                                                                        \\
Learning Rate           & 5e-4                                                                           & 1e-4                                                          & 1e-4                                                         \\
Layer Decay Rate        & 0.5                                                                            & 0.65                                                          & 0.65                                                         \\
Scheduler               & \multicolumn{3}{c}{$\eta = \eta_0 \cos \left( \frac{7\pi k}{16K} \right)$}                                                                                                                                    \\
Model EMA Momentum      & \multicolumn{3}{c}{0.0}                                                                                                                                                                                       \\
Prediction EMA Momentum & \multicolumn{3}{c}{0.999}                                                                                                                                                                                     \\
Weak Augmentation       & \begin{tabular}[c]{@{}c@{}}Random Crop, \\ Random Horizontal Flip\end{tabular} & \multicolumn{2}{c}{\begin{tabular}[c]{@{}c@{}}Random Resized Crop and Interpolation, \\ Random Horizontal Flip\end{tabular}} \\
Strong Augmentation     & \multicolumn{3}{c}{RandAugment~\cite{cubuk2020randaugment}}                                                                                                                             \\
$\alpha_p$              & \multicolumn{3}{c}{1}                                                                                                                                                                                         \\
$\lambda_p$             & \multicolumn{3}{c}{1.0}                                                                                                                                                                                       \\ \hline
\end{tabular}
}
\label{tab:hyperparams_img}
\end{table*}

\begin{table*}[h]
\centering
\caption{Hyperparameters of NLP tasks.}
\begin{tabular}{c|ccc}
\hline
Dataset                 & AG News              & Yahoo! Answer             & Amazon Review            \\ \hline
Max Length              & \multicolumn{3}{c}{512}                                                     \\
Model                   & \multicolumn{3}{c}{Bert-Base}                                               \\
Weight Decay            & \multicolumn{3}{c}{1e-4}                                                    \\
Labeled Batch size      & \multicolumn{3}{c}{4}                                                       \\
Unlabeled Batch size    & \multicolumn{3}{c}{4}                                                       \\
Learning Rate           & 5e-5                 & 1e-4                      & 5e-5                     \\
Layer Decay Rate        & 0.65                 & 0.65                      & 0.75                     \\
Scheduler               & \multicolumn{3}{c}{$\eta = \eta_0 \cos\left(\frac{7\pi k}{16K}\right)$}     \\
Model EMA Momentum      & \multicolumn{3}{c}{0.0}                                                     \\
Prediction EMA Momentum & \multicolumn{3}{c}{0.999}                                                   \\
Weak Augmentation       & \multicolumn{3}{c}{None}                                                    \\
Strong Augmentation     & \multicolumn{3}{c}{Back-Translation~\cite{qizhe2020}} \\
$\alpha_p$              & \multicolumn{3}{c}{1}                                                       \\
$\lambda_p$             & \multicolumn{3}{c}{1.0}                                                     \\ \hline
\end{tabular}
\label{tab:hyperparams_nlp}
\end{table*}

\subsection{Additional implementation details}
\label{sec:imp_detail_baseline}

\paragraph{Pseudo-supervision}
To ensure a fair comparison with SSL baselines, we use the same number of labeled and unlabeled batches participating during the optimization process.
In each optimization step, when a labeled batch of size $B_L$ and an unlabeled batch of size $B_U$ are given, PS optimizes the following loss:
\begin{align*}
\mathcal{L}_{PS} = \frac{1}{B} \left( \sum^{B_L}_{i=1} \mathcal{H}(y_i, \mathbf{p}(y|x_i)) + \sum^{B_U}_{i=1} \mathcal{H}(\hat{y}^U_i, \mathbf{p}(y|u_i)) \right).
\end{align*}
where $B = B_L + B_U$.
Pseudo-supervision has been shown to improve performance in semi-supervised learning scenarios, particularly when a zero-shot method provides accurate pseudo-labels for the task at hand~\cite{hegselmann2023tabllm,nam2023semi}. However, when the zero-shot predictions are inaccurate, it remains unclear whether pseudo-supervision will still be beneficial.

\paragraph{Pseudo-label as feature input}
To implement pseudo-label as feature input baseline, we change the model architecture to accomodate one-hot pseudo-labels.
When training sample $x_i$ with its corresponding one-hot coverted pseudo-label $\tilde{y_i}$ passes through the training model with encoder $g$ and classifier head $h$, the model concatenates input feature output $g(x)$ and one-hot pseudo-label $\tilde{y_i}$ as additional feature output, producing $h(g(x_i),\tilde{y_i})$ as output.
Note that the input dimension of head $h$ is adjusted to accomodate additional feature dimension (number of classes).
With each sample having pair pseudo-label, this modification allows running SSL baseline (Adamatch) without further modification, and model can learn to accomodate or ignore pseudo-label feature depending on how helpful they are in learning the task.

\paragraph{Accomodating embeddings to ZeroMatch}
To accomodate feature embeddings from foundation models in ZeroMatch, we concatenate embeddings with our feature encoder output before passing to classifier head layer.
When the training model has architecture of encoder $g$ and classifier head $h$, we modify the model's MLP head layer's input dimension so that encoded feature and embeddings are concatenated and put into the head $h$.
As a result, when training sample $x_i$ with its embedding $e_i$ are given, the model outputs $h(g(x_i), e_i)$.
Same procedure is applied to the auxiliary KD classifier head for stage 2 in our algorithm, producing $h_p(g(x_i), e_i)$ to be supervised with teacher pseudo-label.

\section{Additional comparsion with previous works}
\label{sec:additional_comparsion}

\subsection{Compare with fine-tuning foundations models for Semi-supervised learning}
\label{sec:additional_comparsion_ft_fm}

We provide comparisons with recent works that fine-tunes the foundation models for SSL.
We compare our apporach with GRIP~\cite{menghini2023enhancing}, CPL~\cite{zhang2024candidate} and FineSSL~\cite{ganerasing} which adopts prompt tuning strategies to vision-language models such as CLIP~\cite{radford2021learning} and iteratively refines pseudo-label predictions based on confidence and features.

While direct comparisons of these works with ours is not possible due to settings largely different in model and training algorithms, we present settings where our method can outperform these methods, and provide comparison on the details on training and  information used.
We benchmark our method on exact same data settings that these works have conducted their experiments on.

\paragraph{Visual Prompt Tuning Methods}
visual prompt tuning method (VPT)~\cite{jia2022visual} is prompt-tuning method for Vision-language models that tunes the prefix part of input layer while keeping textual encoder fixed. FineSSL, CPL and GRIP provide VPT benchmarks with CLIP models.
We provide comparison of ZeroMatch on FineSSL's CIFAR-100 results in Table~\ref{tab:compare_vpt_cifar}, and results of GRIP and CPL with Flowers102 and Resisc45 dataset in Table~\ref{tab:compare_vpt_flower}.

\begin{table*}[ht]
    \centering
    \small
    \caption{
    Comparison with FineSSL's VPT (visual prompt tuning) results on CIFAR100 with 400 labels (4 labels per class). * indicates numbers published in FineSSL.
    }
    \resizebox{\textwidth}{!}{
    \begin{tabular}{c|ccc|c}
    \hline
    Method           & FM Inference info used       & Training backbone & Fine-tuning method          & Accuracy            \\ \hline
    FineSSL*         & -                            & CLIP ViT-B-16     & VPT-deep & 80.44±0.24          \\ \hline
    Zero-shot        & CLIP ViT-L-14 - pseudo-label & -                 & -                           & 60.65±0.00          \\
    ZeroMatch (ours) & CLIP ViT-L-14 - pseudo-label & ViT-S-2           & Full fine-tuning            & \textbf{86.73±1.42} \\ \hline
    \end{tabular}
    }
    \label{tab:compare_vpt_cifar}
\end{table*}

\begin{table*}[ht]
    \centering
    \small
    \caption{
    Comparison with GRIP and CPL's VPT results on Flowers102 and Resisc45 datasets with 2 labels per class. * indicates numbers published in the paper.
    }
    \resizebox{\textwidth}{!}{
    \begin{tabular}{c|ccc|c|c}
    \hline
    Method           & FM Inference info used        & Training backbone & Finetuning method    & Flowers102          & RESISC              \\ \hline
    Zero-shot*       & CLIP ViT-B-32 - pseudo-label  & -                 & -                    & 63.67               & 54.48               \\
    GRIP*            & CLIP text encoder - embedding & CLIP ViT-B-32     & VPT & 67.95±1.2           & 71.22±0.77          \\
    CPL*             & CLIP text encoder - embedding & CLIP ViT-B-32     & VPT & 73.52±0.37          & 78.46±0.74          \\ \hline
    Zero-shot        & GPT-4.1 pseudo-label          & -                 & -                    & 88.37               & 79.28               \\
    ZeroMatch (ours) & GPT-4.1 pseudo-label          & ViT-S-16          & Full fine-tuning     & \textbf{95.17±0.88} & \textbf{87.65±0.73} \\ \hline
    \end{tabular}
    }
    \label{tab:compare_vpt_flower}
\end{table*}

Our results show ZeroMatch's outperforming scores for all the baseline methods.
This is largely due to leveraging high-quality outputs of larger foundation model - For CIFAR-100, we use pseudo-labels from CLIP ViT-L-14, and for Flowers102, we leverage GPT-4.1.
We also note that the training backbone we use is smaller than the baselines. We use ViT-Small which uses less number of attention heads compared with CLIP-ViT-B models.

This result implies that our method can outperform many baselines that fine-tunes the foundation models by simply involving the highest quality foundation model, while training a smaller model for the task on low-compute device.

\paragraph{Textual Prompt Tuning Methods}

Textual prompt tuning ~\cite{zhou2022learning} is another common prompt-tuning based methods which tunes the input prefix of textual encoder of vision-language models, while keeping visual encoder fixed.
Textual encoder is trained with embeddings obtained from large visual encoder, and it can be implemented with using the visual encoder through inference mode only. This setting resembles our problem setup in that the inference outputs of larger models are used to train smaller model.
We provide benchmarks comparing our method to GRIP and CPL's textual prompt tuning results in Table~\ref{tab:compare_tpt_flower}.

\begin{table*}[ht]
    \centering
    \small
    \caption{
    Comparison with GRIP and CPL's TPT (Textual Prompt Tuning) results on Flowers102 and Resisc45 datasets with 2 labels per class. * indicates numbers published in the paper.
    }
    \resizebox{\textwidth}{!}{
    \begin{tabular}{c|ccc|c|c}
    \hline
    Method           & FM Inference info used                           & Training backbone & Finetuning method     & Flowers102          & RESISC              \\ \hline
    Zero-shot*       & CLIP ViT-B-32 - pseudo-label                     & -                 & -                     & 63.67               & 54.48               \\
    GRIP*            & CLIP ViT-B-32 - embedding                        & CLIP text encoder & Textual prompt tuning & 83.6±0.68           & 74.11±0.68          \\
    CPL*             & CLIP ViT-B-32 - embedding                        & CLIP text encoder & Textual prompt tuning & 89.66±0.36          & 80.98±0.11          \\
    Zero-shot*       & CLIP ViT-L-14 - pseudo-label                     & -                 & -                     & 73.98               & 62.67               \\
    CPL*             & CLIP ViT-L-14 - embedding                        & CLIP text encoder & Textual prompt tuning & 96.80±0.63          & 87.75±0.29          \\ \hline
    Zero-shot        & GPT-4.1 pseudo-label                             & -                 & -                     & 88.37               & 79.28               \\
    ZeroMatch (ours) & GPT-4.1 pseudo-label                             & ViT-S-16          & Full fine-tuning      & 95.17±0.88          & 87.65±0.73          \\ \hline
    ZeroMatch (ours) & \begin{tabular}[c]{@{}c@{}}GPT-4.1 pseudo-label + \\ CLIP ViT-L-14 - embedding\end{tabular} & ViT-S-16          & Full fine-tuning      & \textbf{97.20±0.41} & \textbf{88.52±0.77} \\ \hline
    \end{tabular}
    }
    \label{tab:compare_tpt_flower}
\end{table*}

While our method paired with pseudo-labels from GPT-4.1 can outperform CPL and GRIP with CLIP-ViT-B model, CPL's results with CLIP ViT-L outperforms our score.
We additionally benchmark our method with embeddings from CLIP-ViT-L model, which CPL uses as the supervision source.
The results show that our method improves with added embedding information, outperforming other baselines, implying that embedding is a helpful information source for learning the benchmark task.

\subsection{Doubly Robust Self-training}
\label{sec:additional_comparsion_doubly}

Doubly Robust Self-Training~\cite{zhu2024doubly} addresses a problem setting similar to ours, proposing a method that robustly improves performance despite varying quality of pseudo-labels. The method employs the following loss function:

\ifx\isarxiv\undefined
    \begin{align*}
     & \mathcal{L}_{DR}  = \frac{1}{N_L} \sum^{N_L}_{i=1} \mathcal{H}(y_i, \mathbf{p}(y|x_i)) - \frac{1}{N_L} \sum^{N_L}_{i=1} \mathcal{H}(\hat{y}^L_i, \mathbf{p}(y|x_i)) \\
    & + \frac{1}{N} \left( \sum^{N_L}_{i=1} \mathcal{H}(\hat{y}^L_i, \mathbf{p}(y|x_i)) + \sum^{N_U}_{i=1} \mathcal{H}(\hat{y}^U_i, \mathbf{p}(y|u_i)) \right).
    \end{align*}
\else
    \begin{align*}
     \mathcal{L}_{DR}  = \frac{1}{N_L} \sum^{N_L}_{i=1} \mathcal{H}(y_i, \mathbf{p}(y|x_i)) - \frac{1}{N_L} \sum^{N_L}_{i=1} \mathcal{H}(\hat{y}^L_i, \mathbf{p}(y|x_i)) + \frac{1}{N} \left( \sum^{N_L}_{i=1} \mathcal{H}(\hat{y}^L_i, \mathbf{p}(y|x_i)) + \sum^{N_U}_{i=1} \mathcal{H}(\hat{y}^U_i, \mathbf{p}(y|u_i)) \right).
    \end{align*}
\fi

When the pseudo-labels are accurate, the loss function approximates training with the full dataset $\mathcal{D}_L$ and $\mathcal{D}_U$, both with ground truth labels. Conversely, when the pseudo-labels are inaccurate, the loss function asymptotically approaches training with only the labeled samples $\mathcal{D}_L$. The work provides convergence improvement guarantees over training with labeled data alone, with varying pseudo-label quality, under certain regularity conditions.
While the study presents experimental results using a single pseudo-label set inferred from a pre-trained model, its practical effectiveness with inaccurate pseudo-labels in low-label settings is unexplored.
To implement this method and ensure a fair comparison with ours, we use the same number of labeled and unlabeled batches participating during the optimization process.
In each optimization step, when a labeled batch of size $B_L$ and an unlabeled batch of size $B_U$ are given, DR optimizes the following:

\ifx\isarxiv\undefined
    \begin{align*}
     & \mathcal{L}_{DR}  = \frac{\alpha_t}{B_L} \sum^{B_L}_{i=1} \mathcal{H}(y_i, \mathbf{p}(y|x_i)) - \frac{\alpha_t}{B_L} \sum^{B_L}_{i=1} \mathcal{H}(\hat{y}^L_i, \mathbf{p}(y|x_i)) \\
    & + \frac{1}{B} \left( \sum^{B_L}_{i=1} \mathcal{H}(\hat{y}^L_i, \mathbf{p}(y|x_i)) + \sum^{B_U}_{i=1} \mathcal{H}(\hat{y}^U_i, \mathbf{p}(y|u_i)) \right).
    \end{align*}
\else
    \begin{align*}
     \mathcal{L}_{DR}  = \frac{\alpha_t}{B_L} \sum^{B_L}_{i=1} \mathcal{H}(y_i, \mathbf{p}(y|x_i)) - \frac{\alpha_t}{B_L} \sum^{B_L}_{i=1} \mathcal{H}(\hat{y}^L_i, \mathbf{p}(y|x_i)) + \frac{1}{B} \left( \sum^{B_L}_{i=1} \mathcal{H}(\hat{y}^L_i, \mathbf{p}(y|x_i)) + \sum^{B_U}_{i=1} \mathcal{H}(\hat{y}^U_i, \mathbf{p}(y|u_i)) \right).
    \end{align*}
\fi
where $B = B_L + B_U$ and $\alpha_t$ represents the annealing parameter, following the original implementation in~\cite{zhu2024doubly}.

We provide comparison of doubly-robust method with ZeroMatch on Yahoo Answers and Amazon Reviews in Table~\ref{tab:compare_doubly_robust}. Our results show that our method outperforms doubly-robust method all settings.
We hypothesize this is due to asymptotic guarantees of Doubly-robust Self-training method failing to hold in the low-label regime.

\begin{table*}[ht]
    \centering
    \small
    \caption{Comaparison with Doubly-robust self-training method in Yahoo Answers and Amazon Review with pseudo-labels from FLAN-T5 models. Median and standard deviation of accuracy (\%) of 3 different random seeds are reported. Best score among experiments with same pseudo-label set are in bold. `FM' refers to the foundation model used to generate the pseudo-label set.}
    \resizebox{\textwidth}{!}{
    \begin{tabular}{c|c|ccc|ccc}
    \hline
                       & Dataset            & \multicolumn{3}{c|}{Yahoo Answers}                              & \multicolumn{3}{c}{Amazon Review}                               \\ \hline
    FM                 & Label size         & 250                 & 500                 & 2000                & 125                 & 250                 & 1000                \\ \hline
    \multirow{3}{*}{A} & Doubly-robust      & 51.39±12.13         & 44.32±10.91         & 47.9±5.68           & 46.94±0.72          & 47.25±0.63          & 47.76±0.87          \\
                       & Zero-shot          & 66.63±0.00          & 66.63±0.00          & 66.63±0.00          & 52.37±0.00          & 52.37±0.00          & 52.37±0.00          \\
                       & ZeroMatch   (ours) & \textbf{69.78±0.94} & \textbf{70.51±0.19} & \textbf{71.81±0.07} & \textbf{58.83±0.51} & \textbf{59.69±0.60} & \textbf{60.95±0.13} \\ \hline
    \multirow{3}{*}{B} & Doubly-robust      & 24.53±4.84          & 22.52±5.54          & 33.88±2.07          & 35.8±0.13           & 35.9±0.07           & 35.96±0.03          \\
                       & Zero-shot          & 35.29±0.00          & 35.29±0.00          & 35.29±0.00          & 35.7±0.00           & 35.7±0.00           & 35.7±0.00           \\
                       & ZeroMatch   (ours) & \textbf{67.05±0.76} & \textbf{67.13±0.73} & \textbf{69.61±0.22} & \textbf{50.23±1.35} & \textbf{53.78±1.74} & \textbf{56.12±0.64} \\ \hline
    \end{tabular}
    }
    \label{tab:compare_doubly_robust}
\end{table*}

\section{Additional experiments and discussions}
\subsection{Image classification results with CLIP models.}
\label{sec:image_classification_with_clip}
Image classification benchmark results of ZeroMatch with CLIP models are presented in Table~\ref{tab:image_clip_results}.

\begin{table*}[t!]
\centering
\small
\caption{Accuracy (\%) in CIFAR-100, Flowers102, Resisc45 with pseudo-labels from CLIP models. Median and standard deviation of 3 different random seeds are reported. Best score among experiments with same pseudo-label set are in bold. `FM' refers to the foundation model used to generate the pseudo-label set. }
\resizebox{\textwidth}{!}{
\begin{tabular}{c|c|ccc|c|c}
\hline
                            & Dataset                & \multicolumn{3}{c|}{CIFAR100}                                   & Flowers102          & Resisc45            \\ \hline
FM                          & Label size             & 100                 & 200                 & 400                 & 204                 & 90                  \\ \hline
None                        & Adamatch               & 71.43±2.48          & 78.32±0.46          & 84.02±0.74          & 86.71±0.69          & 78.87±0.80          \\ \hline
\multirow{4}{*}{CLIP-Large} & Zero-shot              & 60.65±0.00          & 60.65±0.00          & 60.65±0.00          & 72.13±0.00          & 60.32±0.00          \\
                            & Pseudo-supervise       & 70.97±0.41          & 71.86±0.63          & 74.05±0.42          & 74.19±0.57          & 63.93±0.21          \\
                            & PL feature input & 71.21±2.56          & 80.11±0.49          & 84.42±0.22          & \textbf{89.51±0.42} & 80.85±1.43          \\
                            & ZeroMatch   (ours)     & \textbf{84.30±0.71} & \textbf{85.77±0.05} & \textbf{86.13±0.01} & 88.94±1.66          & \textbf{83.91±1.54} \\ \hline
\multirow{4}{*}{CLIP-Base}  & Zero-shot              & 48.07±0.00          & 48.07±0.00          & 48.07±0.00          & 60.17±0.00          & 49.63±0.00          \\
                            & Pseudo-supervise       & 62.19±0.26          & 65.96±0.37          & 71.03±0.23          & 64.09±0.30          & 55.82±0.84          \\
                            & PL feature input & 71.3±3.26           & 78.85±1.45          & \textbf{83.88±0.73} & 87.56±0.29          & 80.85±2.26          \\
                            & ZeroMatch   (ours)     & \textbf{79.78±0.99} & \textbf{80.87±0.62} & 82.69±0.23          & \textbf{89.61±2.39} & \textbf{82.58±0.85} \\ \hline
\end{tabular}
}
\label{tab:image_clip_results}{}
\end{table*}

\subsection{Ablation study}
\label{sec:ablation_study}

Two main components that constructs our method (other than SSL training) is separate knowledge disillation (KD) stage, and learning KD as auxiliary loss for SSL's objective. We validate effectiveness of each component through following ablation study.

\textcircled{1} \textit{Is auxiliary KD loss necessary?}
To validate this, we compare our method with the case without auxiliary KD loss, which essentially runs plain AdaMatch algorithm after running KD in stage 1. We name this apporach ZeroMatch without auxliary loss (`ZM w.o. aux.' in tables) and provide benchmark result in Table~\ref{tab:compare_wo_aux}.
Our results show that adding auxiliary KD loss helps, especially in the low-label setting. For example, In 125-label setting of Amazon Review with GPT-4o, our original method improves plain AdaMatch by 2.22\%. Since pseudo-labels inferred from labeled data can become inaccurate in low-label setting, adding addtional supervision with teacher pseudo-labels can be particularly helpful in this case.

\textcircled{2} \textit{Does having separate KD stage necessary?}
Since stage 2 of our algorithm also includes auxiliary KD loss, one may wonder if running stage 2 without stage 1 in our algorithm is enough to achieve the goal of merging benefits of KD and SSL.
To clear this out, we compare our method with the case without running stage 1 knowledge distillation. We name this apporach ZeroMatch without stage 1 (`ZM w.o stage 1' in tables) and provide benchmark result in Table~\ref{tab:compare_zm_stage2}.
Our results show that while ZeroMatch without stage 1 can sometimes achieve the similar performance to original ZeroMatch (ex. in Yahoo Answers) it may also produce large gap in performance depending on the dataset. For example, in 250-label setting Aamzon Review with `A' pseudo-label set, ZM stage 2 only reveals 3.94\% accuracy drop compared to original.

\begin{table*}[ht]
    \centering
    \small
    \caption{Comaparison with ZeroMatch without auxliary loss in Yahoo Answers and Amazon Review with pseudo-labels from GPT-4o and LLama3.3 models. Median and standard deviation of accuracy (\%) of 3 different random seeds are reported. Best score among experiments with same pseudo-label set are in bold. `FM' refers to the foundation model used to generate the pseudo-label set.}
    \resizebox{\textwidth}{!}{
    \begin{tabular}{c|c|ccc|ccc}
    \hline
                                  & Dataset            & \multicolumn{3}{c|}{Yahoo Answers}                              & \multicolumn{3}{c}{Amazon Review}                               \\ \hline
    FM                            & Label size         & 250                 & 500                 & 2000                & 125                 & 250                 & 1000                \\ \hline
    \multirow{2}{*}{GPT-4o}       & ZM w.o. aux.        & 68.22±0.32          & 69.23±0.53          & 71.45±0.23          & 57.89±0.47          & 59.60±0.33          & 60.16±0.34          \\
                                  & ZeroMatch   (ours) & \textbf{70.90±1.07} & \textbf{71.11±0.09} & \textbf{72.09±0.32} & \textbf{60.12±0.33} & \textbf{59.82±0.39} & \textbf{60.19±0.48} \\ \hline
    \multirow{2}{*}{LLama3.3-70B} & ZM w.o. aux.        & 69.30±0.53          & 70.54±0.54          & 71.54±0.31          & 56.59±1.78          & 59.04±0.21          & \textbf{60.13±0.08} \\
                                  & ZeroMatch   (ours) & \textbf{71.28±0.29} & \textbf{71.46±0.95} & \textbf{71.68±0.19} & \textbf{57.36±0.81} & \textbf{59.46±0.13} & 60.06±0.28          \\ \hline
    \end{tabular}
    }
    \label{tab:compare_wo_aux}
\end{table*}

\begin{table*}[ht]
    \centering
    \small
    \caption{Comaparison with ZeroMatch Stage 2 only in Yahoo Answers and Amazon Review with pseudo-labels from FLAN-T5 models. Median and standard deviation of accuracy (\%) of 3 different random seeds are reported. Best score among experiments with same pseudo-label set are in bold. `FM' refers to the foundation model used to generate the pseudo-label set.}
    \resizebox{\textwidth}{!}{
    \begin{tabular}{c|c|ccc|ccc}
    \hline
                       & Dataset            & \multicolumn{3}{c|}{Yahoo Answers}                              & \multicolumn{3}{c}{Amazon Review}                               \\ \hline
    FM                 & Label size         & 250                 & 500                 & 2000                & 125                 & 250                 & 1000                \\ \hline
    \multirow{2}{*}{A} & ZM w.o stage 1    & \textbf{70.01±0.63} & 69.76±0.22          & 71.22±0.27          & 55.16±0.30          & 55.75±0.65          & 58.53±0.36          \\
                       & ZeroMatch   (ours) & 69.78±0.94          & \textbf{70.51±0.19} & \textbf{71.81±0.07} & \textbf{58.83±0.51} & \textbf{59.69±0.60} & \textbf{60.95±0.13} \\ \hline
    \multirow{2}{*}{B} & ZM w.o stage 1    & 66.6±0.13           & 67.05±0.44          & 69.22±0.53          & 45.53±4.91          & 53.19±1.68          & \textbf{56.9±0.63}  \\
                       & ZeroMatch   (ours) & \textbf{67.05±0.76} & \textbf{67.13±0.73} & \textbf{69.61±0.22} & \textbf{50.23±1.35} & \textbf{53.78±1.74} & 56.12±0.64          \\ \hline
    \end{tabular}
    }
    \label{tab:compare_zm_stage2}
\end{table*}

\subsection{Hyperparameter sensitivity analysis}
\label{sec:hyperparameter_sensitivity}

For the simiplicity of setting, we mainly benchmark our method with hyperparameters $\alpha_p=1$ and $\lambda_p=1$ without finding the optimal setting.
The reasoning behind this choice to make KD loss same as fully-supervised training when pseudo-label is accurate, and enabling annealing to allow more freedom to accomodate labeled data in the early stage of AdaMatch, which develops initial confident predictions for unlabeled data.

In this paragraph, we run a sensitivity analysis on combinations of these parameters.
we conduct experiments with four different sets: $(\alpha_p, \lambda_p) = (0, 0.5), (0, 1.0), (1, 0.5), (1, 1.0)$. The results with NLP datasets with FLAN-T5 models are shown in Table~\ref{tab:hp_sensitivity}.
We find that while $(\alpha_p, \lambda_p) = (1, 1.0)$ can be a good candidate on some datasets (ex. Yahoo Answers), there may exist other optimal parameters that can further improve performance.

\begin{table*}[ht]
    \centering
    \small
    \caption{Hyperparameter sensitivity results in Yahoo Answers (250 labels) and Amazon Review (125 labels) with pseudo-labels from FLAN-T5 models.}
    \begin{tabular}{c|cc}
    \hline
    $\alpha, \lambda$ & Yahoo Answers       & Amazon Review       \\ \hline
    (0, 0.5)        & 69.15±0.46          & 59.55±1.69          \\
    (0, 1.0)        & 69.90±0.87          & \textbf{60.20±1.02} \\
    (1, 0.5)        & 68.26±0.94          & 59.31±1.43          \\
    (1, 1.0)        & \textbf{70.71±0.36} & 59.47±1.26          \\ \hline
    \end{tabular}
    \label{tab:hp_sensitivity}
\end{table*}

\subsection{Computational cost analysis}
\label{sec:compute_cost}

We provide memory usage and computation cost associated with our method and SSL baseline(AdaMatch) for CIFAR100 dataset in Table~\ref{tab:compute_cost}.
The usages are measured in our machine with 1 NVIDIA A10G GPU.
We note that our method consumes around 50\% more training time compared with AdaMatch. This is due to having separate knoweledge distillation at stage 1, with utilizing half of the batch size of AdaMatch.
Also our method uses slightly more VRAM due to back-propagating through weak-augmented samples in KD auxiliary loss in stage 2, while in AdaMatch weak-augmented samples are only used with forward pass to get predictions and back-propagation happens in strong augmented samples.
Some options to reduce the compute usage are applying early stopping in stage 1 knowledge distillation when pseudo-label prediction accuracy is above certain threshold, and involving strong augmentation in KD auxiliary loss instead of weak augmented sample so that backpropagation does not happen for weak augmented samples.

\begin{table*}[h]
    \centering
    \small
    \caption{Computational cost on training with CIFAR100.}
    \begin{tabular}{c|cc}
    \hline
    Method    & VRAM(GB) & Training time (sec) \\ \hline
    Adamatch  & 4.2      & 41100               \\ \hline
    ZeroMatch & 5.4      & 61662               \\ \hline
    \end{tabular}
    \label{tab:compute_cost}
\end{table*}

\section{Limitations}
\label{sec:limitations}
In this section, we discuss few limitations of our work.
\begin{itemize}
\item We only benchamrk our method on the case where fixed number of labeled data are sampled per each class. This may not demonstrate practical scenario where labels and classes have imbalanced distributions. Future directions can include benchmarking our method on such setting.
\item While most of our experimental results show that our method outperforms zero-shot method, there is no methodological guarantee this always hold. Future directions can explore ensuring robust improvement over zero-shot.
\item We also note that our method's performance inherently depends on the quality of the pseudo-labels generated from foundation models. Robustness against labels that are not from foundation models (e.g. weak supervision signal, patterned noise) are not tested and may not work with our method. Future direction can further explore application to different type of label noise.
\end{itemize}

\ifx\isarxiv\undefined
    \section{Broader Impact}
    \label{sec:broader_impact}

    Our work aims to make semi-supervised learning more accessible by reducing the computational resources needed to leverage foundation models. This has positive implications for democratizing AI development, particularly benefiting researchers and organizations with limited computing infrastructure.

    However, improved semi-supervised learning techniques could also lower the barrier for developing AI systems with potential misuse cases. While our method focuses on improving efficiency rather than expanding capabilities, we acknowledge the need for careful consideration of deployment contexts.

    The adaptive weighting mechanism we introduce could potentially mitigate some biases in foundation models by automatically reducing reliance on unreliable predictions. However, this requires further investigation, particularly regarding the method's behavior across different demographic groups and task domains.

\else
\fi


\end{document}